\pdfoutput=1
\documentclass[11pt]{article}

\usepackage[]{acl}
\usepackage{enumitem}

\usepackage{times}
\usepackage{latexsym}
\usepackage{xcolor}
\usepackage{caption}
\RequirePackage[color=yellow!20,linecolor=red,textsize=scriptsize, disable]{todonotes}
\usepackage[T1]{fontenc}
\usepackage[utf8]{inputenc}

\usepackage{microtype}

\usepackage{color, colortbl}
\definecolor{Gray}{gray}{0.9}
\usepackage{multirow,adjustbox}
\usepackage{diagbox}
\newcolumntype{M}[1]{>{\centering\arraybackslash}m{#1}}

\usepackage{bm,amsmath,amssymb,fancyhdr,graphicx}
\usepackage{hyperref}
\usepackage{graphicx}
\usepackage{subcaption}

\DeclareMathOperator{\llm}{LLM}

\usepackage[ruled,vlined]{algorithm2e}
\def\ztitle{Language Models can Exploit Cross-Task In-context Learning for Data-Scarce Novel Tasks}
\title{\ztitle}

\author{Anwoy Chatterjee$^{*}$, \hspace{0.2mm} Eshaan Tanwar$^{*}$, \hspace{0.2mm} Subhabrata Dutta, \hspace{0.2mm} Tanmoy Chakraborty\\
Indian Institute of Technology Delhi \\ 
\small {
 \texttt{anwoy.chatterjee@ee.iitd.ac.in}, \:
 \texttt{eshaantanwar2000@gmail.com}
 }\\
 \small {\texttt{subha0009@gmail.com}, \:
 \texttt{tanchak@iitd.ac.in}
 }}
 
\begin{document}
\maketitle
\def\thefootnote{*}\footnotetext{Equal contribution named in alphabetical order}\def\thefootnote{\arabic{footnote}}
\begin{abstract}
Large Language Models (LLMs) have transformed NLP with their remarkable In-context Learning (ICL) capabilities. Automated assistants based on LLMs are gaining popularity; however, adapting them to novel tasks is still challenging. While colossal models excel in zero-shot performance, their computational demands limit widespread use, and smaller language models struggle without context. This paper investigates whether LLMs can generalize from labeled examples of predefined tasks to novel tasks. 
Drawing inspiration from biological neurons and the mechanistic interpretation of the Transformer architecture, we explore the potential for information sharing across tasks. 
We design a cross-task prompting setup with three LLMs and show that LLMs achieve significant performance improvements despite no examples from the target task in the context. Cross-task prompting leads to a remarkable performance boost of 107\% for LLaMA-2 7B, 18.6\% for LLaMA-2 13B, and 3.2\% for GPT 3.5 on average {\color{black} over zero-shot prompting, and performs comparable to standard in-context learning}. The effectiveness of generating pseudo-labels for in-task examples is demonstrated, and our analyses reveal a strong correlation between the effect of cross-task examples and model activation similarities in source and target input tokens. This paper offers a first-of-its-kind exploration of LLMs' ability to solve novel tasks based on contextual signals from different task examples.
\end{abstract}

\section{Introduction}
\label{sec:intro}

Large Language Models (LLMs) have revolutionized the state of Natural Language Processing for the past few years. With the ability of In-Context Learning (ICL), one can adopt an LLM to almost any task without costly gradient updates \citep{DBLP:journals/corr/abs-2005-14165}. At the same time, automated assistants, built on top of foundational LLMs, are being popularized~\citep{pahune2023several}. A crucial challenge with this escalating usage popularity is handling novel tasks. Humongous models like GPT-4 are able to deliver up-to-par performance even in a zero-shot regime~\citep{openai2023gpt4}. However, the computational requirements of deploying such large-scale models counteract the practicality of their usage en masse. Relatively smaller LMs, on the other hand, suffer drastically in the absence of in-context examples. {\color{black} The availability of labeled examples usually varies across the use cases of the language model. For example, in an NLP research setup, expert users can quickly come up with a few handwritten examples. However, when we consider mass-scale usage of average users who are not experienced prompt engineers or need quick answers, the zero-shot performance of a model becomes extremely crucial. For example, assuming the popular usage of ChatGPT, very few non-expert users would opt to write down examples while asking ChatGPT to perform some tasks.}

This naturally raises the question of whether one can make an LLM generalize from labeled examples of a predefined set of tasks to an input defining a novel task. In the world of biological neurons, such abilities are commonplace: inculcating specific limb usage into an untrained limb while training the opposite limb~\cite{cross-education}, or relatively easier adoption of newer skills from the culminated experience of older skills~\cite{neural-task,marton2021efficient}. Drawing a blunt parallel between biological neurons and LLMs would be naive. 
However, one can find a supporting intuition in the mechanistic interpretation of the Transformer architecture~\cite{transformer-circuits,automatic-circuit,interpretability-in-the-wild}.
\if 0
, given that ICL abilities in LLMs manifest with special neural circuitries, called {\em induction heads}~\cite{induction-heads}. Induction heads typically look for patterns similar to the target input within the context examples and copy information pathways from context residual streams to target residual streams. 
\fi 
One can argue that if information pathways necessary to solve a novel task are similar to those corresponding to some different task from a task library, an LLM may gather useful information across tasks. Earlier evidence found by \citet{tanwar-etal-2023-multilingual} also elicits intuitive motivation as they showed that LLMs can learn to infer from cross-lingual examples if proper alignment is provided.

\begin{figure*}[!t]
\begin{center}
\includegraphics[width=\textwidth]{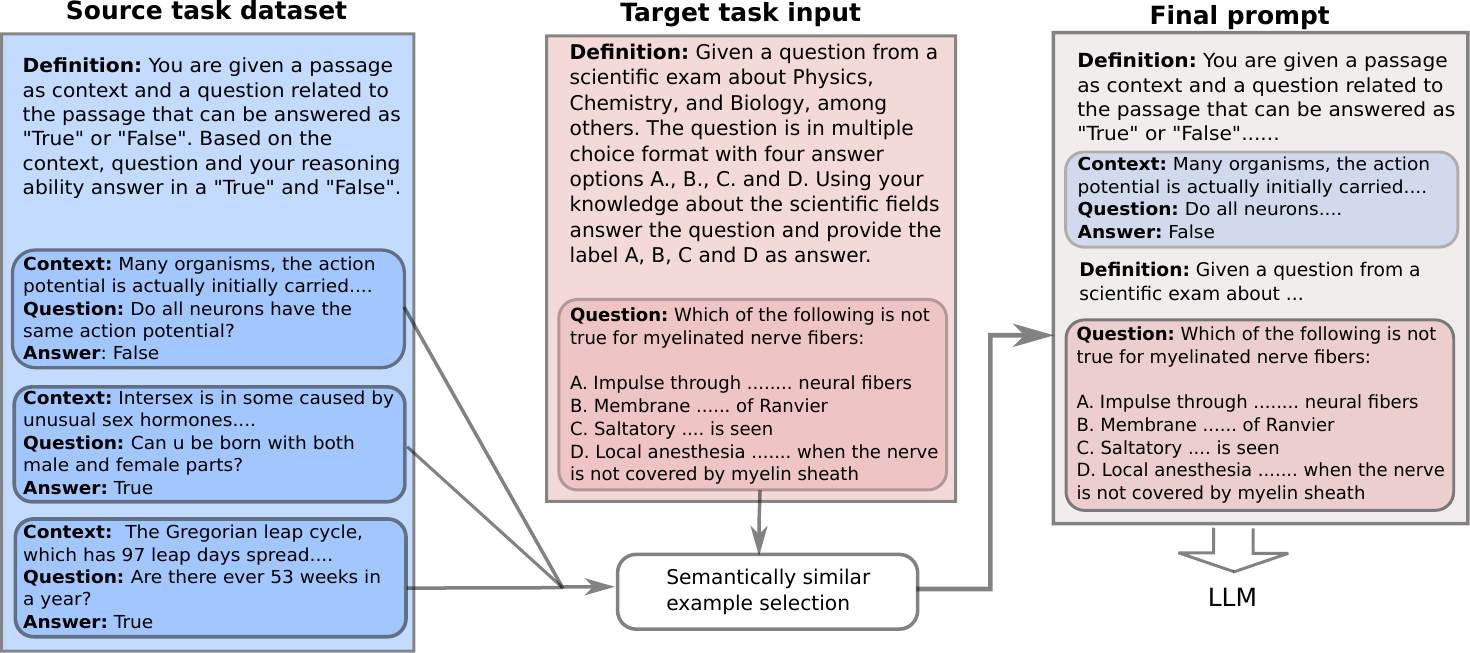}
\caption{In this working example, we aim to solve a question from MedMCQA using demonstrations from BoolQ. To do so, we sample a semantically similar demonstration from the source task and then use this demonstration along with task descriptors of the source and target tasks to generate a cross-task prompt that is fed to an LLM.}
\label{fig:change_in_k_psudo_CTICL}
\end{center}
\vspace{-5mm}
\end{figure*}

\if 0
However, existing studies on mechanistic interpretation suggest that the neural circuits within LLMs are much more complex in organization~\cite{induction-heads,interpretability-in-the-wild}. It would be extremely difficult to identify such circuit sharing in a cross-task setting. Instead, we look for an empirical proof-of-concept about the viability of such a phenomenon.
\fi
In this work\footnote{Code available at \href{https://github.com/C-anwoy/Cross-Task-ICL.git}{https://github.com/C-anwoy/Cross-Task-ICL}}, we design a cross-task prompting setup (Section~\ref{sec:cross-task-method}) using three different LLMs: LLaMA-2 7B and 13B, and GPT 3.5; we select \(50\) different pairs of tasks where one serves as a {\em source} (i.e., context example task) and the other as {\em target}. Despite no examples from the target task presented in the context, LLMs can produce a staggering improvement over the zero-shot regime; on average, cross-task prompting improves performance by \(107\%\) for LLaMA-2 7B, \(18.6\%\) for LLaMA-2 13B and \(3.2\%\) for GPT 3.5 (Section~\ref{sec:cross-task-main-result}). With multiple source tasks, cross-task performance is even better than, if not comparable, usual in-task prompting ({\bf Contribution \#1}). 
However, learning from examples of different tasks is heavily sensitive to the choice of source task for a given target and the LLM is prone to copy the label space of the source task into the target. To circumvent this, we propose a pseudo-labeling based approach: in a data-scarce setup, cross-task prompting with majority voting is first employed to generate noisy, in-task examples; these are subsequently used for standard few-shot prompting ({\bf Contribution \#2}).
Finally, we provide introductory analysis towards interpreting cross-task signal transfer by dissecting the model activations. We find that the cross-task signal transfer is abrupt and happens at later layers, with the effective layers widely varying for different target tasks  ({\bf Contribution \#3}).
In a nutshell, this is the first exploration of LLMs' ability to learn to solve novel tasks based on the contextual signals of different task examples.

%\todo{I dont like the color combination of Fig 1}

% what do we mean by low-resource 
% \begin{enumerate}
%     \item Low resource
%     \item Domain Specific
%     \item Tough tasks
% \end{enumerate}

% \textcolor{red}{TBD}

\section{Prompting techniques}
\label{sec:cross-task-method}

{\bf Task definition} for a given task is a natural language instruction for the LM describing what is asked of it (see Figure~\ref{fig:change_in_k_psudo_CTICL}). Since a cross-task setup would result in in-context examples from different tasks (with different label spaces), such a definition is necessary to discriminate.

Next, given two datasets \(D_s\) and \(D_t\) corresponding to two different tasks with task definitions \(d_s\) and \(d_t\), respectively, we formalize the cross-task prompting as inferring the output \(\hat{y}_t\) for an input \(x_t\in D_t\) conditioned upon a demonstration from dataset \(D_s\):
\begin{equation*}
\small
    \hat{y}_t = \underset{y}{\mathrm{argmax}}\, p(y|(d_s \oplus x_s \oplus y_s \oplus d_t \oplus x_t))
\end{equation*}
% \subsection{Preliminaries}
In this setup, we denote \(D_s\) and \(D_t\) as source and target task datasets, respectively. Note that the formalization is equivalent to 1-shot prompting where the only provided example input-output pair comes from a different dataset. Table \ref{tab:cross-task-prompts} of Appendix \ref{sec:appendix} presents illustrative examples of prompts in this setup. Our experiments suggest that \(>\)1-shot setup in cross-task prompting does not improve (and often deteriorates) performance. 

\citet{liu-etal-2022-makes} showed that for a target input $x_t$, generating context $C$ from semantically similar examples leads to not only better results but also a more robust method of prompting. Similarly, \citet{tanwar-etal-2023-multilingual} showed that ICL could be done in a cross-lingual setup by aligning the semantic and task signals of the source and target languages.

Drawing inspiration from these works, we set up the cross-task prompting regime with sampling examples $x_s$ from the source task dataset $D_s$ that are semantically similar to the target input $x_t$.
To extract semantically similar examples, we first utilize Sentence-BERT \cite{reimers-2019-sentence-bert} to extract the sentence embedding of target and source inputs~\footnote{\textcolor{black}{We also experimented with E5 and LLaMA-2 7B last layer outputs, see Appendix~\ref{sec:semantic-context-creation}}.}. Following this, based on the cosine similarity between the embeddings, we select top source examples. 
% \textcolor{red}{We experimented with several models for extracting semantically similar $C$, analysis of that can be found in Appendix~\ref{sec:semantic-context-creation}}.
% todo{we have context, input, labels. What is exemplar?}
% Algorithm \ref{alg:sim-prompt} shows a concise implementation of the method.

{\bf In-task examples combined with cross-task prompting. }
\label{sec:In-task prompting combined with cross-task prompting}
So far, we have assumed the unavailability of any labeled target dataset. But, if we do have a labeled target example, could prepending a source task boost its prompting performance?

To emulate such a scenario,
%\todo{dont write anotators; otherwise people will be confused. we are not doing any annotation} 
a labeled example from the target dataset $T_t$ is sampled. This 
 labeled example $(x_{lt},y_{lt}),\ \text{where}\ x_{lt} \in T_t$, is then used to construct prompt. This mixed setup can be formalized as,
\begin{equation*}
\small
    \hat{y}_t = \underset{y}{\mathrm{argmax}}\, p(y|(d_s \oplus x_s \oplus y_s \oplus d_t \oplus x_{lt} \oplus y_{lt} \oplus x_t))
\end{equation*}
 
Examples of prompts are provided in Table \ref{tab:icl-with-cross-task} of Appendix \ref{sec:appendix}.

\section{Results and Analysis}
\label{sec:results}

\begin{table*}[!t]
\centering
\adjustbox{max width=1\linewidth}{
\begin{tabular}{lccccccccccc}
\hline
 \backslashbox{TAR}{SRC}& Zero-shot & ARC-Easy & AG-news & BoolQ & Com-QA & C-POS & C-NER & MNLI & QQP  & RACE  & SST2 \\ \hline
\multicolumn{12}{c}{\cellcolor[HTML]{C0C0C0}LLaMA-2 7B}                                                                                \\ \hline
ARC-Challenge        & 4.6       & \textbf{43.6}     & 33.6    & 35.0  & \textbf{43.6}           & 33.8          & 34.2          & 34.2 & 36.4 & 42.8  & 33.2 \\
Financial-Phrasebank & 34.1      & 43.5     & 62.1    & 40.9  & 14.6           & 1.4           & 0.4           & 62.7 & 44.7 & 53.4  & \textbf{65.0} \\
MedMCQA              & 4.2       & 31.4     & 26.8    & 28.0  & \textbf{33.0}           & 26.2          & 24.0          & 23.0 & 26.0 & 31.6  & 23.2 \\
SciQ                 & 8.0       & 59.0     & 45.4    & 49.0  & \textbf{65.6}           & 34.4          & 29.7          & 44.8 & 25.4 & 64.4  & 39.0 \\
Social-i-QA          & 41.1      & 44.3     & 40.1    & 40.3  & 48.5           & 38.9          & 39.1          & 38.9 & 39.7 & \textbf{49.1}  & 39.3 \\ \hline
Average              & 18.4      & 44.3     & 41.6    & 38.6  & 40.71          & 26.9          & 25.5          & 40.7 & 34.4 & \textbf{48.3} & 39.9 \\ \hline
\multicolumn{12}{c}{\cellcolor[HTML]{C0C0C0}LLaMA-2 13B}                                                                               \\ \hline
ARC-Challenge        & 52.0      & \textbf{59.2}     & 50.6    & 54.6  & 57.4           & 50.6          & 49.8          & 49.2 & 52.8 & 56.8  & 50.8 \\
Financial-Phrasebank & 65.4      & 66.6     & 78.6    & 65.8  & 61.7           & 2.2           & 13.17         & 79.0 & \textbf{82.4} & 66.5  & 72.8 \\
MedMCQA              & 9.2       & 37.2     & 34.2    & 36.6  & 38.8           & 31.6          & 32.6          & 30.2 & 30.6 & \textbf{39.0}  & 31.4 \\
SciQ                 & 55.8      & \textbf{83.4}     & 76.2    & 80.2  & \textbf{83.4}           & 76.4          & 78.6          & 77.2 & 71.2 & 82.8  & 72.6 \\
Social-i-QA          & 55.3      & 60.8     & 55.8    & 56.5  & 63.5           & 56.8          & 55.9          & 53.5 & 55.7 & \textbf{63.7}  & 53.1 \\ \hline
Average              & 47.5      & 61.4     & 59.1    & 58.7  & 60.9           & 43.5          & 46.0          & 57.8 & 58.5 & \textbf{61.7} & 56.1 \\ \hline
\multicolumn{12}{c}{\cellcolor[HTML]{C0C0C0}GPT3.5}                                                                               \\ \hline
ARC-Challenge        & 74.6      & 77.2     & 74.4    & 77.8  & 76.2           & 70.2          & 70.8          & 75.2 & 74.0 & \textbf{78.2}  & 72.6 \\
Financial-Phrasebank & 57.5      & 79.8     & \textbf{83.6}    & 78.6  & 82.0           & 50.3          & 64.3          & 72.2 & 74.2 & 73.4  & 75.6 \\
MedMCQA              & 49.6      & 49.8     & 48.5    & \textbf{50.0}  & 48.0           & 47.4          & 45.0          & 46.0 & 49.0 & 47.6  & 48.0 \\
SciQ                 & 91.2      & 91.4     & 89.9    & 90.6  & 91           & 87.8          & 82.0          & 88.8 & 89.8 & \textbf{92.2}  & 89.0 \\
Social-i-QA          & 76.0      & 76.2     & 75.8    & 74.0  & 75.2           & 73.2          & 74.0          & 75.8 & 73.0 & \textbf{77.2}  & 74.8 \\ \hline
Average              & 69.8      & \textbf{74.9}     & 74.5    & 74.2  & 74.5           & 65.8          & 67.2          & 71.6 & 72.0 & 73.7 & 72.0 \\ \hline
\end{tabular}
}
\caption{Accuracy for cross-task setup using one source example. Source tasks are mentioned in columns, and target tasks are in rows. Cross-task prompting brings improvement over zero-shot prompting for certain source-target pairs. (Abbreviations: Com-QA \(\rightarrow\) Commonsense-QA, C-POS \(\rightarrow\) Conll2003-POS, C-NER \(\rightarrow\) Conll2003-NER).}
\label{tab:cross-task}
\vspace{-5mm}
\end{table*}

%\todo{Captions for Tables 1 and 2 do not clearly communicate what is show. Please elaborate properly.}
\noindent \textbf{Datasets and experimental setup.}
Our corpus of tasks consists of ten source and five target tasks. We consider ARC-Easy \cite{Clark2018ThinkYH}, AG-news \cite{Zhang2015CharacterlevelCN}, BoolQ \cite{clark-etal-2019-boolq}, Commonsense-QA \cite{talmor-etal-2019-commonsenseqa}, Conll2003-POS \cite{tjong-kim-sang-de-meulder-2003-introduction}, Conll2003-NER \cite{tjong-kim-sang-de-meulder-2003-introduction}, MNLI \cite{N18-1101}, QQP \cite{DBLP:journals/corr/abs-1907-01041}, RACE \cite{lai-etal-2017-race} and SST2 \cite{socher-etal-2013-recursive} to be our source tasks. {\color{black} Our motivation remains to incorporate domain diversity and difficulty of the problems while choosing target tasks to emulate the ``novel task'' phenomenon as closely as possible since learning from cross-task examples makes sense only when the target task is truly data-scarce.
Our target tasks are Social-i-QA \cite{DBLP:journals/corr/abs-1904-09728}, SciQ \cite{SciQA2023}, MedMCQA \cite{pmlr-v174-pal22a}, Financial-Phrasebank \cite{Malo2014GoodDO}, and ARC-Challenge \cite{Clark2018ThinkYH}; first four of these require domain expertise, while the last one is a more challenging version of the one present in the source tasks}. In total, this gives us $50$ unique cross-task setups (see Appendix \ref{sec:appendix-dataset} for dataset details. Table \ref{tab:source-task-definitions} and Table \ref{tab:target-task-definitions} of Appendix \ref{sec:appendix} show task definitions corresponding to source and target tasks, respectively). The dataset size is standardised by sampling $10,000$ and $500$ examples for each source and target task, i.e., $|D_s|=10,000$ and $|D_t|=500$. For our experiments, we used the 7-billion and 13-billion variants of LLaMA-2 \cite{touvron2023llama}, and text-davinci-003 \cite{DBLP:journals/corr/abs-2005-14165} referred to as GPT3.5. \textcolor{black}{We experiment with greedy and force decoding setup and selected greedy decoding as our standard for all experiments (see Appendix \ref{sec:force-decoding} for more details).} We set the number of examples used to create cross-task context as one for all our experiments, unless mentioned explicitly.

\noindent \textbf{Does cross-task prompting work?}
\label{sec:cross-task-main-result}
As evident from Table \ref{tab:cross-task}, cross-task prompting significantly improves performance compared to zero-shot prompting ({\color{black} see Table \ref{tab:significance-test} in Appendix for results of significance testing}). The best overall source-target pair improves performance by 162\% for LLaMA-2 7B, 30\% for LLaMA-2 13B and 7\% for GPT3.5. 
% {\color{red} We confirm the statistical significance of the improvements by conducting a one-tailed T-test for each source task across all samples of the target tasks, the results of which are presented in Table \ref{tab:significance-test} of Appendix}. 
We note that different models give the best performance in different source-target pairs and not all coupling of source-target tasks seem to work; e.g., Commonsense-QA as a source task decreases performance on Financial-Phrasebank, but is the best source task for ARC-Challenge and MedMCQA (for LLaMA-2 7B). Token classification tasks (POS and NER) seem to depreciate performance for LLaMA-2 13B and GPT3.5; their performance for LLaMA-2 7B is also sub-par compared to other source-target pairings. RACE and ARC-Easy are robust source tasks that improve performance for all target tasks in all models. ARC-Easy, MNLI and BoolQ can also be considered to have this robustness to some degree, as they only hurt the performance in one or two cases.

On average, cross-task prompting improves performance by 107\% for LLaMA-2 7B, 18.6\% for LLaMA-2 13B and 3.2\% for GPT3.5. This is a strong argument for prompting in a cross-task manner when we lack labeled target data, especially using small models, which have poor zero-shot abilities ~\citep{wei2022chain}. We also experiment with LLaMA-2 7B Chat model to analyse the performance of cross-task prompting for instruction-tuned models (see Appendix \ref{sec:instruction-tuned}).

% \todo{All appendix materials should be cited from the main text}

% \todo{same table labels in multiple tables}

\noindent \textbf{Performance with dissimilar source tasks.} Few of the chosen source tasks have slight overlap in their broad topic or format with some of the target tasks, for example, source datasets like ARC-Easy, Commonsense-QA, RACE consists of questions in multiple-choice format, similar to target datasets like ARC-Challenge, Social-i-QA, SciQ and MedMCQA. To analyse the efficacy of cross-task prompting with completely dissimilar source tasks, let us consider only the source tasks whose topic and format are absolutely disjoint from all target tasks. The following five source tasks can be considered dissimilar to the target tasks -- BoolQ, Conll2003-POS, Conll2003-NER, MNLI and QQP. As reported in Table \ref{tab:cross-task}, even with these dissimilar source tasks, we observe a substantial performance improvement of 80.5\% for LLaMA-2 7B, 11.4\% for LLaMA-2 13B, and 0.5\% for GPT3.5 on average over zero-shot prompting. Moreover, if we consider only BoolQ, MNLI and QQP, the average performance gains become 105.9\% for LLaMA-2 7B, 22.8\% for LLaMA-2 13B, and 4\% for GPT3.5. This shows that cross-task prompting can especially enable the relatively smaller LMs to  substantially boost their zero-shot performance even using tasks dissimilar to the target.

\noindent \textbf{Importance of source task definitions.} To further investigate the role of task definitions in cross-task prompting, we check the effect of removing source task definitions on performance. 
We note an average drop of 11\% for LLaMA-2 7B and 8\% for LLaMA-2 13B when we prompt without using source task definitions (see Table \ref{tab:no_source_inst} in Appendix). Hence, definitions play a crucial role in cross-task prompting.

\noindent \textbf{Increasing number of examples for cross-task prompting.}
Unlike in-task prompting \cite{DBLP:journals/corr/abs-2005-14165}, where increasing the number of examples increases prompting performance, in cross-task prompting, performance does not improve with an increase in source examples  (c.f. Fig \ref{fig:k_graph}). For most target tasks, increasing source examples does not affect performance, while for some, the performance decreases.

\noindent \textbf{Semantically similar vs random example selection.}
\label{sec:Semantically similar vs randomly demonstrations prompts}
As shown in Table \ref{tab:rand_vs_sim}, in a cross-task setup, choosing the examples randomly leads to substantially poorer performance than when we generate the context $C$ with semantically similar examples; concerning is the fact that, in many cases, it causes 0\% accuracy. This may be caused by the model getting confused without any semantic alignment in the prompt. {\color{black} Furthermore, random labeling of in-context examples \citep{symbol-tune} result in near-random performance (see Table~\ref{tab:random-label-icl} in Appendix~\ref{sec:random-label-icl}).}

\begin{table}[!t]
\centering
\adjustbox{max width=1\linewidth}{
\begin{tabular}{lccccc}
\hline
\backslashbox{TAR}{SRC} & AG-news      & BoolQ         & Com-QA & \multicolumn{1}{l}{MNLI} & \multicolumn{1}{l}{QQP} \\ \hline
\multicolumn{6}{c}{\cellcolor[HTML]{C0C0C0}Random}                                                                        \\ \hline
ARC-Challenge        & $26.6_{3.0}$ & $3.7_{0.7}$   & $0.0_{0.0}$    & $29.5_{3.7}$             & $26.8_{2.9}$            \\
Financial-Phrasebank & $0_{0}$      & $1.4_{1.3}$   & $13.9_{14.75}$ & $0.0_{0.0}$              & $0.0_{0.0}$             \\
MedMCQA              & $26.1_{1.5}$ & $20.8_{1.1}$  & $0.0_{0.0}$    & $25.1_{0.9}$             & $25.0_{0.2}$            \\
SciQ                 & $34.5_{3.2}$ & $10.0_{4.25}$ & $0.0_{0.0}$    & $34.0_{4.1}$             & $28.7_{0.8}$            \\
Social-i-QA          & $36.8_{1.3}$ & $0.6_{0.1}$   & $0.0_{0.0}$    & $24.9_{2.7}$             & $29.27_{3.9}$           \\ \hline
\multicolumn{6}{c}{\cellcolor[HTML]{C0C0C0}Semantically Similar}                                                            \\ \hline
ARC-Challenge        & 36.6         & 35.0          & 43.6           & 34.2                     & 36.4                    \\
Financial-Phrasebank  & 62.1         & 40.9          & 14.6           & 62.7                     & 44.7                    \\
MedMCQA              & 26.8         & 28.0          & 33.0           & 23.0                     & 26.0                    \\
SciQ                 & 45.4         & 49.0          & 65.6           & 44.8                     & 25.4                    \\
Social-i-QA          & 40.1         & 40.3          & 48.5           & 38.9                     & 39.7       \\ \hline            
\end{tabular}
}
\caption{Accuracy when cross-task examples are picked randomly vs when they are pickled based on semantic similarity. Experiments were done on the LLaMA-2 7B model. For random prompting, we generated the results over three seeds and reported the average performance (Com-QA: Commonsense-QA).}
\label{tab:rand_vs_sim}
\vspace{-1mm}
\end{table}

\begin{table}[!t]
\centering
\adjustbox{max width=1\linewidth}{
\begin{tabular}{lccc}
\hline
                                         & Best source cross-task                  & Random mixed cross-task & Best mixed cross-task \\ \hline
\multicolumn{4}{c}{\cellcolor[HTML]{C0C0C0}LLaMA-2 7B}                                                               \\ \hline
ARC-Challenge                            &        $51.6$                &     $42.8$             &          $44.0$                \\
Financial-Phrasebank                     &        $64.7$          &     $48.3$            &          $56.0$   \\
MedMCQA                                  &         $34$          &     $28.6$            &          $29.0$                  \\
SciQ                                     &        $46.2$          &     $60.5$             &          $67.2$            \\ 
Social-i-QA                              &        $42.7$          &     $42.9$            &          $42.9$                \\ \hline
Average                                  &       $47.84$          &     $44.62$               &           $47.82$                \\ \hline
\multicolumn{4}{c}{\cellcolor[HTML]{C0C0C0}LLaMA-2 13B}                                                               \\ \hline
ARC-Challenge                            &        $66.4$                &     $57.2$             &          $60.4$                \\
Financial-Phrasebank                     &        $76.6$          &     $71.4$            &          $61.5$   \\
MedMCQA                                  &         $38.4$          &     $38.3$            &          $41.2$                  \\
SciQ                                     &        $84.6$          &     $80.5$             &          $84.2$            \\ 
Social-i-QA                              &        $49.3$          &     $61.9$            &          $61.27$                \\ \hline
Average                                  &       $63.06$          &     $61.86$               &           $61.71$                \\ \hline
\end{tabular}
}
\caption{Accuracy in mixed cross-task prompting. Four examples are used to create context $C$ in all setups.}
\label{tab:mixed_prompting}
\vspace{-3mm}
\end{table}

\begin{table*}[!t]
\centering
\adjustbox{max width=1\linewidth}{
\begin{tabular}{lccccccccccc}
\hline
\backslashbox{TAR}{SRC} & In-task  & +ARC-Easy & +AG-news & +BoolQ & +Com-QA & +C-POS & +C-NER & +MNLI & +QQP  & +RACE  & +SST2 \\ \hline
\multicolumn{12}{c}{\cellcolor[HTML]{C0C0C0}LLaMA-2 7B}                                                                                \\ \hline
ARC-Challenge        & 49.8 & \textbf{51.2}     & 41.6    & 45.8  & 49.6           & 43.6          & 40.2          & 41.6 & 45.2 & 50.0  & 42.8 \\
Financial-Phrasebank & 33.3 & 33.3     & 33.3    & 33.3  & 33.5           & 33.3          & 34.5          & 35.7 & 33.3 & 34.5  & \textbf{50.3} \\
MedMCQA              & 32.4 & 33.6     & 29.2    & 30.6  & 32.8           & 31.0         & 30.8          & 28.2 & 30.2 & \textbf{34.2}  & 31.2 \\
SciQ                 & 67.4 & \textbf{73.4}     & 60.4    & 68.6  & 71.6           & 59.0          & 57.4          & 58.6 & 57.9 & 71.2  & 62.4 \\
Social-i-QA          & 43.3 & 49.9     & 46.7    & 44.5  & \textbf{58.3}          & 44.7          & 44.3          & 41.5 & 43.9 & 51.7  & 43.1 \\ \hline
Average              & 45.2 & 48.3     & 42.2    & 44.5  & 48.1           & 42.3          & 41.4          & 41.1 & 42.1 & \textbf{48.3} & 45.9 \\ \hline
\multicolumn{12}{c}{\cellcolor[HTML]{C0C0C0}LLaMA-2 13B}                                                                               \\ \hline
ARC-Challenge        & 61.8 & \textbf{63.0}     & 60.8    & 61.6  & 61.0           & 61.2          & 60.0          & 60.0 & 59.6 & 61.2  & 58.8 \\
Financial-Phrasebank & \textbf{86.0} & 77.0     & 71.2    & 77.6  & 82.4           & 65.1          & 69.4          & 75.2 & 64.8 & 81.8  & 85.6 \\
MedMCQA              & 40.2 & 40.8     & 37.6    & 40.4  & 41.0           & 38.2          & 39.0          & 38.6 & 38.6 & \textbf{41.8}  & 37.2 \\
SciQ                 & 84.8 & \textbf{85.8}     & 83.0    & 83.8  & 85.6           & 81.4          & 81.4          & 80.2 & 79.2 & 84.0  & 79.6 \\
Social-i-QA          & 54.8 & \textbf{62.1}     & 52.09   & 52.3  & 60.9           & 53.3          & 53.7          & 54.1 & 53.5 & 61.6  & 53.9 \\ \hline
Average              & 65.5 & 65.7     & 60.9    & 63.1  & 66.2           & 59.8          & 60.7          & 61.6 & 59.1 & \textbf{66.0}  & 63.0 \\ \hline
\multicolumn{12}{c}{\cellcolor[HTML]{C0C0C0}GPT3.5}                                                                               \\ \hline
ARC-Challenge        & 78.6      & 78.2     & 78.2    & 78.8  & \textbf{79.8}           & 77.4          & 78.4          & 78.4 & 78.4 & 77.6  & 79.0 \\
Financial-Phrasebank & 75.4      & 80.2     & 79.4    & 79.6  & 83.8           & 70.4          & 80.2          & 80.2 & 81.0 & \textbf{81.6}  & 77.6 \\
MedMCQA              & \textbf{52.0}      & 51.2     & 48.6    & 51.2  & 47.4           & 48.2          & 49.4          & 49.4 & 48.8 & 49.8  & 49.8 \\
SciQ                 & 92.6      & 91.2     & 91.8    & 92.6  & 92.4           & 91.8          & 92.0          & 92.0 & 91.4 & \textbf{93.0}  & 91.6 \\
Social-i-QA          & 75.6      & 77.0     & 76.2    & 74.4  & 76.6           & 76.2          & \textbf{77.4}          & 77.4 & 75.0 & 77.0  & 75.6 \\ \hline
Average              & 74.8      & 75.6     & 74.8    & 75.3  & \textbf{76.0}           & 72.8          & 74.2          & 75.0 & 74.9 & 75.8 & 74.7 \\ \hline
\end{tabular}
}
\caption{Accuracy for in-task combined with cross-task prompting setup; prompt context is created using one source example and one human-labelled target example. Cross-task prompting compliments in-task promoting, leading to better performance (Com-QA: Commonsense-QA, C-POS: Conll2003-POS, C-NER: Conll2003-NER).}
\label{tab:cross-task+ICL}
\vspace{-1mm}
\end{table*}

\noindent \textbf{Mixed cross-task prompting.}
So far, we have seen that cross-task prompting works for single source-target task pairs. Next, we experiment on using multiple source tasks to construct the prompt context. To explore such a setup, we prompt LLaMA models using three methods:

\begin{enumerate}[leftmargin=*,noitemsep,topsep=0pt]
    \item \textbf{Best source cross-task:} We select the best source task for every target task using Table \ref{tab:cross-task} and sample four semantically similar examples from that source task.
    \item \textbf{Random mixed cross-task:} To see if a diverse set of tasks is beneficial, we randomly sample four source tasks and construct the prompt using most semantically similar examples.
    \item \textbf{Best mixed cross-task:} This method is a combination of the first two; we use the top four best source tasks from Table \ref{tab:cross-task} and sample a semantically similar source example from each task.
\end{enumerate}

 Table  \ref{tab:mixed_prompting} shows that a mixed prompting mechanism does not perform better than the ``best source cross-task'' prompting method. On the contrary, it seems to hurt the performance of the model; in fact, single-source task prompting with only one example (Table \ref{tab:cross-task}) seems to do better than mixed prompting. Diversity seems to lead the model to get more confused, thus hurting its performance.

\noindent \textbf{Combining in-task with cross-task prompting.} Combining cross-task prompting with labeled target examples improves performance, as seen in Table \ref{tab:cross-task+ICL}; apart from Financial-Phrasebank in LLaMA-2 13B and MedMCQA in GPT3.5, for all other instances there exist a source task that improves the performance when coupled with the in-task example. However, we see that this improvement is immensely dependent on the source-target task pair chosen and unlike Table \ref{tab:cross-task}, we are unable to find robust source datasets that improve the performance throughout the setup. Nevertheless, in multiple source-target instances, there is a noteworthy improvement. 
% In Table~\ref{tab:in-task-comparison}, we compare the outcomes of best mixed task prompting (from Table~\ref{tab:cross-task+ICL}) vs. 1-shot and 2-shot in-task prompting. We can see that, even 2-shot in-task prompting can often deteriorate performance from 1-shot setup while mixed task prompting is able to deliver comparable or better performance. Typically, with relative smaller models like the LLaMA-2 series, betterment with mixed setup is much more profound.

\begin{figure*}[!t]
% \begin{center}
\begin{minipage}[t]{0.75\textwidth}
\vspace{0pt}
\includegraphics[width=\linewidth]{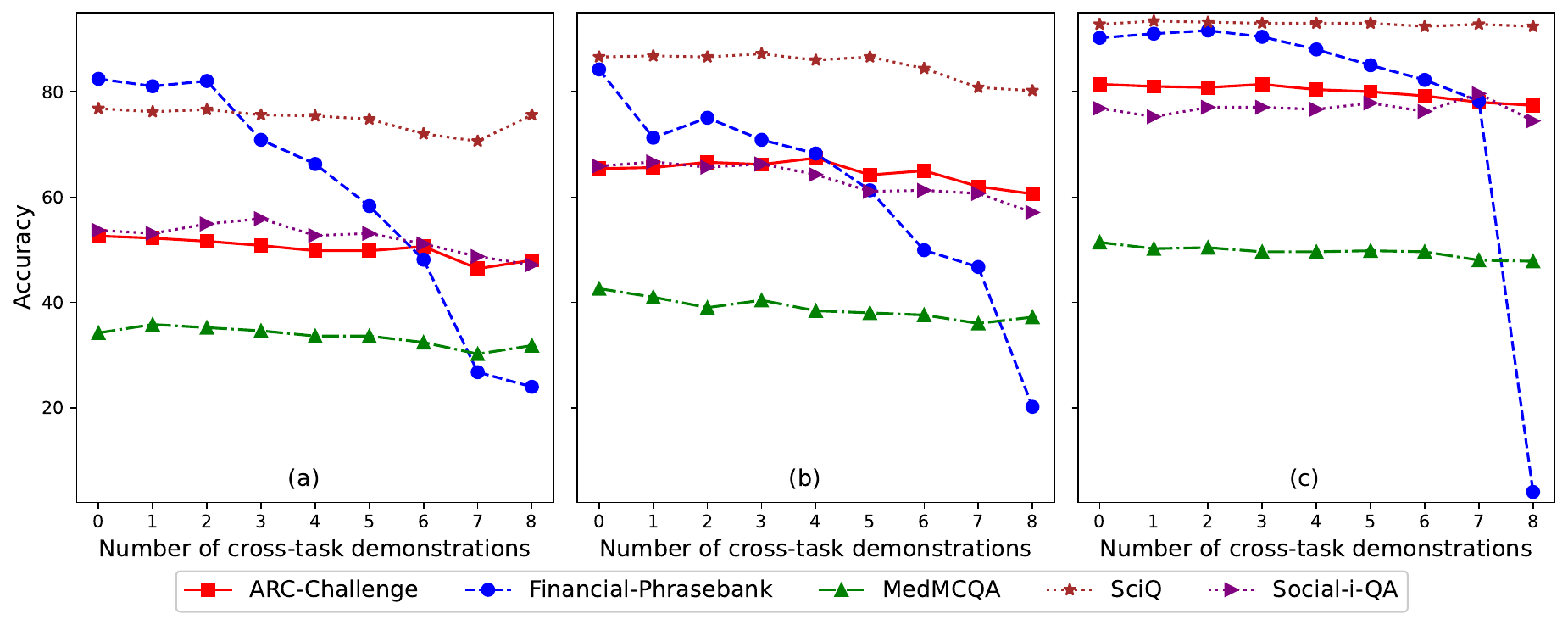}
\end{minipage}\hfill
\begin{minipage}[t]{0.23\textwidth}
\vspace{0pt}
\captionsetup{labelfont={small, color=black},font={small, color=black}}
\caption{Variation of accuracy with the number of cross-task demonstrations in the context C for 8-shot prompt, where C consists of a mixture of cross-task and in-task examples, using (a) LLaMA-2 7B, (b) LLaMA-2 13B and (c) GPT3.5. Except for Financial-Phrasebank, an increase in cross-task examples give a stable performance.}
\label{fig:mixed_context}
\end{minipage}
% \end{center}
% \vspace{-7mm}
\end{figure*}

{\color{black}To study the interaction between heterogeneous tasks in the context, we experiment by varying the number of source tasks and target task demonstrations in the context. 
% In this setup, to create a $C$ from $k$-shots ,we use $m$ source-tasks, all giving one semantically similar demonstrations each and $k-m$ semantically similar target-task demonstration, i.e. $C$ will be :
% \begin{equation*}
% \small
%     \begin{split}
%     C = d_{s_m} \oplus x_{s_m} \oplus y_{s_m} \oplus ... \oplus d_{s_1} \oplus x_{s_1} \oplus y_{s_1} \oplus \\
%      d_t \oplus x_{k-m} \oplus y_{k-m} \oplus ... \oplus x_{1} \oplus y_{1} \oplus x_t
%     \end{split}
% \end{equation*}
Figure~\ref{fig:mixed_context} shows the variation of accuracy for 8-shots, as we gradually move from an entirely in-task context to a complete cross-task context. We observe that for all target tasks, apart from Financial-Phrasebank, the accuracy with an 8-shot in-task prompt and an 8-shot cross-task prompt (with \textit{best-mixed cross-task} strategy) is almost the same in all three LLMs.}

\section{Pseudo-label generation using cross-task prompting}
\label{sec:Pseudo Label generation cross-task prompting}

% Our $T_t$ are domain-specific and tough for the LLMs to solve. In such a scenario annotation of a dataset requires expert annotators which becomes resource-intensive. 
% \citet{DBLP:journals/corr/abs-2005-14165} showed that providing an LLM with a few labelled examples from the target task to create context $C$ leads to substantial improvement in performance compared to zero-shot prompting. 
% Earlier studies \cite{10.1162/tacl_a_00492,DBLP:journals/corr/abs-2109-06270} utilised pseudo-labeling of unlabeled data to improve LLMs' prompting abilities.

Thus far we observe that cross-task prompting, though sometimes capable of even outperforming standard in-task prompting, is particularly sensitive to the choice of source task. Furthermore, the performance does not scale with the number of source task examples provided. Drawing inspiration from earlier works on pseudo-labeled examples to construct prompts~\citep{10.1162/tacl_a_00492,DBLP:journals/corr/abs-2109-06270}, we propose a more practical method for potential usage of cross-task prompting.

Given a small unlabeled dataset $D_{pl}\subset D_t$, we assign a pseudo-label to the example using cross-task prompting. This is done using all source tasks available to us, and then a final $y_{pl}$ is assigned to $x_{pl}\in D_{pl}$ based on a majority vote from all the generated answers. Finally, this pseudo-labeled dataset $D_{pl}$ is used to construct the prompt context in an in-context prompting setup. 
% Algorithm \ref{alg:pseudo-dem} shows the step-by-step implementation of this method.

% \begin{algorithm}[!t]\small
% \SetAlgoLined
% \caption{Pseudo label generation using cross-task prompting}\label{alg:pseudo-dem}
% \textbf{Input: }Target input $x_t$, source data $D_s$, task definitions $d_t$ and $d_s$, an unlabeled dataset $D_{pl}$ , and number of samples to extract $k$.

% \textbf{Procedure: }
% $\mathbf{y_{pl}}\gets(x)$\\
% \For{$x_{pl}\in D_{pl}$}{
%     \For{$T_s  \in T$}{
%     $y_{pl}^s\gets x_{pl}\ \text{using Algo}\ \ref{alg:sim-prompt}$
%     }
%     $y_{pl}\gets \maj(y_{pl}^s)$}
% Select $k x$ from $D_{pl}$

% $C\gets d_s\oplus x_{pl}^1\oplus y_{pl}^1\oplus \cdots x_{pl}^k\oplus y_{pl}^k$

% $y_{t}=\llm(y|C\oplus [sep]\oplus x_t)$
% \end{algorithm}

\noindent \textbf{Cross task generated pseudo-examples vs gold label.}
To see the efficacy of our proposed method, we utilise a $D_{pl}$ of size 8 and have three setups to create examples for context: 1. \textit{Gold-label: } We use the annotated version of $D_{pl}$ in the context; 2. \textit{Pseudo-label (ZS): } We use zero-shot prompting to label $D_{pl}$ for the context examples; 3. \textit{Pseudo-label (CT): } We use the cross-task method as proposed in Section \ref{sec:Pseudo Label generation cross-task prompting}.

The results of this experiment are shown in Table \ref{tab:pseudo_label}. For LLaMA models, as evident, pseudo labels generated via cross-task prompting are substantially better than zero-shot pseudo labels; they are of higher quality and lead to comparable performance as Gold-label. 

In the case of GPT 3.5, the scale of improvement is much smaller though. Interestingly, with only two datasets, namely, SciQ and Social-i-QA, we observe the pseudo examples from cross-task prompting to perform worse than gold-label examples, in all three models. Given the comparable performance of cross-task prompt-generated pseudo examples, we expect this to become a viable alternative to traditional ICL in data-scarce scenarios.
% This observation is valid across all the datasets. These results show a better alternative for utilising
% These results provide us with a good alternative to utilise data
% unlabeled target inputs for prompting, in LLaMA models\todo{rephrase this line}. GPT3.5, on the other hand, seems to give a similar performance for all three setups.

\begin{table}[!t]
\centering
\adjustbox{max width=1\linewidth}{
\begin{tabular}{lccc}
\hline
                                         & Gold-label                  & Pseudo-examples(ZS) & Pseudo-examples(CT) \\ \hline
\multicolumn{4}{c}{\cellcolor[HTML]{C0C0C0}LLaMA-2 7B}                                                               \\ \hline
ARC-Challenge                            &       $44.4_{0.37}$          &     $7.2_{4.52}$             &          $44.4_{0.37}$                \\
Financial-Phrasebank                     &       $48.4_{9.56}$          &     $56.5_{2.01}$            &          $56.8_{2.37}$   \\
MedMCQA                                  &       $30.6_{0.84}$          &     $21.0_{14.61}$            &          $30.6_{1.03}$                  \\
SciQ                                     &       $67.7_{0.65}$          &     $3.3_{1.88}$             &          $64.6_{1.51}$            \\ 
Social-i-QA                              &       $46.8_{1.69}$          &     $46.7_{1.57}$            &          $46.4_{1.94}$                \\ \hline
Average                                  &       $47.6_{2.62}$          &     $26.9_{4.91}$               &           $48.5_{1.44}$                \\ \hline
\multicolumn{4}{c}{\cellcolor[HTML]{C0C0C0}LLaMA-2 13B}                                                               \\ \hline
ARC-Challenge                            &       $56.8_{0.47}$          &     $59.7_{0.84}$             &          $58.4_{1.79}$                \\
Financial-Phrasebank                     &       $67.9_{5.03}$          &     $67.9_{5.03}$            &          $67.99_{5.03}$   \\
MedMCQA                                  &       $38.2_{1.31}$          &     $2.6_{1.8}$            &          $39.4_{0.37}$                  \\
SciQ                                     &       $84.1_{0.75}$          &     $29.3_{39.5}$             &          $84.1_{0.75}$            \\ 
Social-i-QA                              &       $63.6_{2.78}$          &     $43.6_{27.7}$            &          $61.4_{3.66}$                \\ \hline
Average                                  &       $62.1_{2.06}$           &    $40.62_{14.97}$                    & $62.2_{2.32}$                         \\ \hline
\multicolumn{4}{c}{\cellcolor[HTML]{C0C0C0}GPT3.5}                                                               \\ \hline
ARC-Challenge                            &       $76.7_{0.89}$          &     $77.4_{1.13}$             &          $77.46_{0.94}$                \\
Financial-Phrasebank                     &       $80.4_{4.35}$          &     $86.2_{2.84}$            &          $83.43_{6.41}$   \\
MedMCQA                                 &       $43.2_{5.65}$          &     $43.4_{3.30}$            &          $47.8_{2.97}$                  \\
SciQ                                     &       $92.8_{1.04}$          &     $91.2_{0.56}$             &          $91.5_{0.25}$            \\ 
Social-i-QA                             &       $77.8_{1.06}$          &     $76.6_{0.65}$            &          $76.6_{0.43}$                \\ \hline
Average                                  &          $74.2_{2.59}$       &     $74.9_{1.69}$                    &   $75.3_{2.2}$                       \\ \hline
\end{tabular}
}
\caption{Accuracy in pseudo-label generated prompting (ZS: zero-shot, CT: cross-task). Pseudo-label (CT) performs comparably to gold-label examples.}
\label{tab:pseudo_label}
\vspace{-5mm}
\end{table}

\begin{figure}[!t]
\begin{center}
\includegraphics[width=\linewidth]{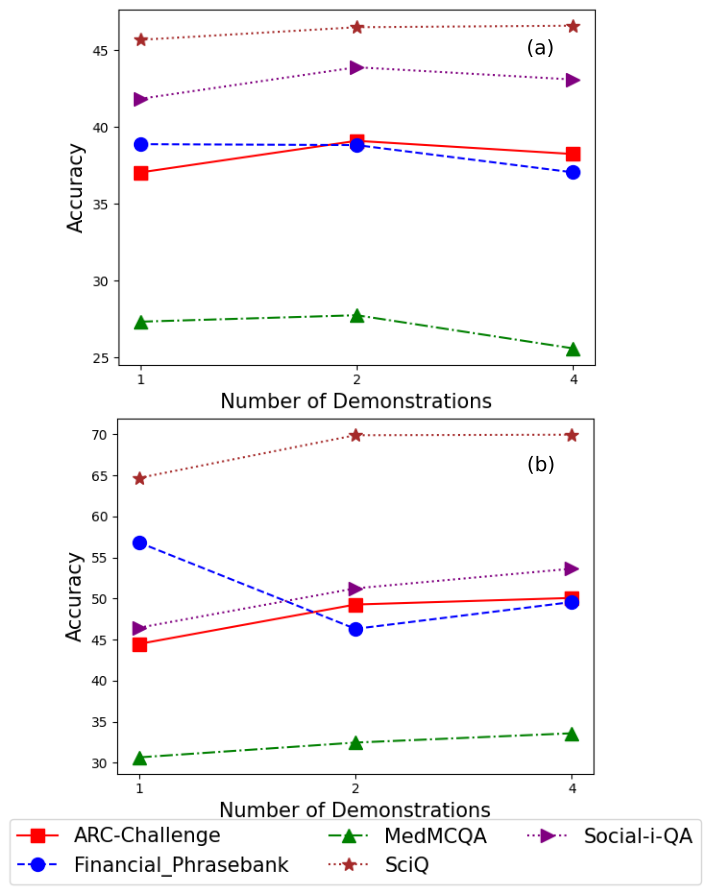}
\caption{ Variation of accuracy with change in the number of -- (a) source demonstrations from $D_{s}$ (the performance largely stagnates at $k=1$), (b) pseudo-label sampled from $D_{pl}$ (the performance seems to increase with an increase in $k$). All experiments are done on LLaMA-2 7B model.}
\label{fig:k_graph}
\end{center}
\vspace{-7mm}
\end{figure}
  % Figure is changed please check and remove comment if no change required
  
\noindent \textbf{Increasing number of pseudo-examples for in-task prompting.} Figure \ref{fig:k_graph} shows the relation between the number of pseudo-demonstrations and accuracy. The general trend shows a rise in performance when more demonstrations are used in creating context $C$, except for Financial-Phrasebank, whose performance decreases with an increase in demonstrations\todo{why? can we justify}.
\begin{figure*}[!h]
    \centering
    \includegraphics[width=\linewidth]{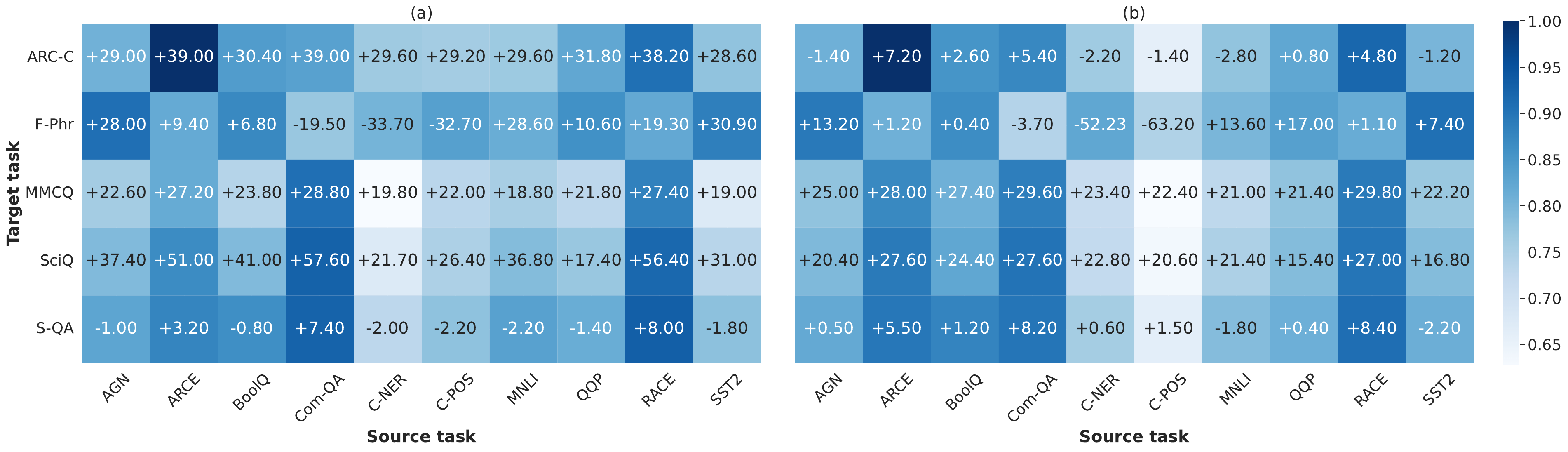}
    \caption{Heat map of cosine similarities between the mean activations of the task definitions for source-target pairs, extracted from the final layer of (a) LLaMA-2 7B (b) LLaMA-2 13B. The values in the cells show the absolute performance improvement over zero-shot prompting. For a target task, the source-target pair with the highest cosine similarity shows the most improvement in 80\% cases.
    (ARC-C: ARC-Challenge, F-Phr: Financial-Phrasebank, MMCQ: MedMCQA, S-QA: Social-i-QA, AGN: AG-news, ARCE: ARC-Easy, Com-QA: Commonsense-QA, C-POS: Conll2003-POS, C-NER: Conll2003-NER). }
    \label{fig:heatmap}
    \vspace{-2mm}
\end{figure*}
\section{Interpretability analysis}

%%% Following is my attempt to shorten the section please review %%%
\textcolor{black}{In this section, we focus on the internal workings of the LLMs using the hidden states of the model and build an understanding of why and how cross-task prompting works.}

\noindent \textcolor{black}{\textbf{What is an ideal source task?} Intuitively thinking of a source task that is similar to the target will result in better transfer of generalised signal from context to target inputs. To test this, we first compare the final layer outputs of the model corresponding to the source and target task definitions, \(d_s\) and \(d_t\), respectively. Figure~\ref{fig:heatmap} shows the cosine similarity between the final layer hidden states corresponding to \(d_s\) and \(d_t\). We note that for 80\% of the time, the source definition that is the most semantically similar to the target definition also serves as its best suitor, leading to the best performance in Table \ref{tab:cross-task}. }

% It is intuitive that source tasks that are similar to the target will result in better transfer of generalizable signal from context to target inputs. To test this, we first compare the final layer outputs of the model corresponding to the source and target task definitions, \(d_s\) and \(d_t\), respectively. Given the final layer activation of the LLM corresponding to a token \(x\) represented as \({\bf h_x}\), we compute two dense representations of the task definitions as follows:
% \begin{equation*}
%         {\bf v}_s = \mathop{\mathbb{E}}_{x\in d_s}[h_x];\: {\bf v}_t = \mathop{\mathbb{E}}_{x\in d_t}[h_x].
% \end{equation*}
% We denote the similarity between two tasks as the cosine similarity between \({\bf v}_s\) and \({\bf v}_t\). As we can see in Figure~\ref{fig:heatmap}, a higher similarity score between the source and the target tasks usually elicits betterment with cross-task prompting. We note that for 80\% of the time the source definition that is the most semantically similar to the target definition also serves as its best suitor, i.e. leads to the best performance in Table \ref{tab:cross-task}. 
% Coupling a target task with its most identical source task leads to an ideal pairing.

\noindent \textcolor{black}{\textbf{What does internal activation indicate? }}To get an even more in-depth picture, we analyse the layerwise activations for the LLaMA-2 7B model. For each layer, we compute the cosine similarities between the mean activations corresponding to the source and target task definitions. For a given target task, we then compute the rank correlation (in terms of Spearman correlation coefficient, Pearson coefficient, and, Kendall's tau) between the cosine similarities and the absolute point performance change from zero-shot prediction for each source task (see Table~\ref{tab:layer_correlation} in Appendix). Precisely, this gives us an approximate idea about the underlying mechanism
of cross-task signal transfer. For all the tasks, we can see a U-shaped pattern in the correlation values \textcolor{black}{(c.f. Fig~\ref{fig:corr_plot})} -- at the starting layers, there is a high correlation between the cross-task activation similarity and the cross-task improvement (likely due to semantically similar example selection), that quickly drops in the middle layers, and increases again in the later layers (only exception being the MedMCQA target task). One can intuitively claim that in the initial layers of the model (Layers 2 to 5), there is more task-specific computation going on where the cross-task transfer of information is the least. For Financial-Phrasebank, these task-specific layers cover more than 80\% of the layers, with a gradual increase in correlation observable only after Layer 28. \citet{icl-task-vectors} provided a similar finding that the influence of the context kicks in only after a certain number of layers via {\em task vectors}. We see that the exact layer after which cross-task demonstrations will start signal to transfer is much more dependent on the target task and can vary widely.

%%%%%%% Corr figure %%%%%%%%%%%%%%
%\if 0
\begin{figure}[!t]
\begin{center}
\includegraphics[width=\linewidth]{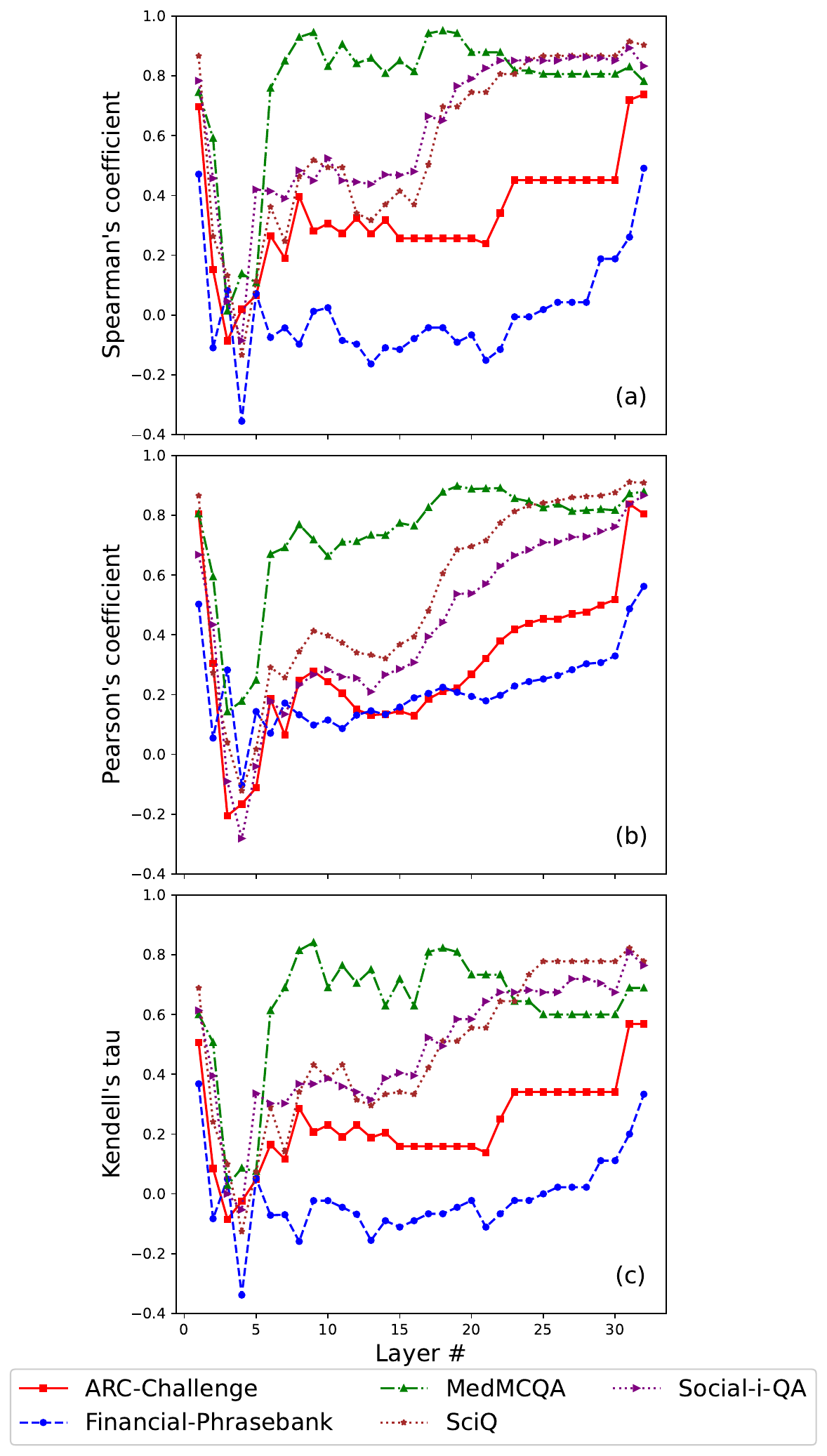}
%\captionsetup{labelfont={color=blue},font={color=blue}}
\caption{ Variation of (a) Spearman's correlation coefficient, (b) Pearson correlation coefficient and (c) Kendall’s tau, with layers as reported in Table \ref{tab:layer_correlation} for LLaMA-2 7B.}
\label{fig:corr_plot}
\end{center}
\vspace{-5mm}
\end{figure}
%\fi

%%%% Error Analysis %%%%
\begin{table*}[!t]
\tiny
\begin{center}
\scalebox{1}{
\begin{tabular}{ |m{0.1cm}|m{10cm}| m{2.5cm}|}
 \hline
 \#&{\bf Prompt}&{\bf Output} \\
  \hline
1.&{\bf Definition:} Given a sentence do token classification by doing Part-of-speech (POS) tagging, ... categories.

{\bf Sentence:} Net profit 6.08 vs 3.98 
{\bf Label:} JJ NN CD NNP CD

{\bf Definition:} Given a sentence ... label the sentence as "negative", "positive" or "neutral"

{\bf Sentence:} Consolidated net sales increased 16 \% ... to a loss of EUR0 .7 m in the prior year period. {\bf Label:}  & \textcolor{red}{NNP CD NNP CD} \\
\hline
2.&
{\bf Definition:} Given a sentence do token classification ... till the entity ends.

{\bf Sentence:} Total shares to be offered 2.5 million 
{\bf Label:} O O O O O O O

{\bf Definition:} Given a sentence ... label the sentence as "negative", "positive" or "neutral"

{\bf Sentence:} The offer of some 30 million shares ... expected to be completed by Oct. 9 , Outokumpu said. 
{\bf Label:} & \textcolor{red}{N N N N N N N} \\
\hline
3.&{\bf Definition:} Given a question answering task ... most appropriate choice that answers the question

{\bf Question:} Which object in our solar system reflects light and is a satellite that orbits around one planet?
A. Earth
...
D. the Moon 
{\bf Answer:} D

{\bf Definition:} Given a question answering task ... most appropriate choice that answers the question

{\bf Question:} Which of the following sets contains only objects that shine as a result of reflected light?

A. moons, planets, and comets
...
D. planets, stars, and moons 
{\bf Answer:} & \textcolor{red}{D} \\
\hline
4.&{\bf Definition:} Given a context and a question ... answer in a "True" and "False".

{\bf Context:} A blacklight (or often black light), also referred to as a UV-A light, ... much visible light. 

{\bf Question:} is a black light and an ultraviolet light the same thing  
{\bf Answer:} True

{\bf Definition:} Given a question from a scientific exam ... provide the label "A", "B", "C" and "D" as answer

{\bf Question:} When exposed to ultraviolet, ... characteristic visible wavelengths, a process called this?

A. pigment
...
D. fluorescence 
{\bf Answer:} & \textcolor{red}{fluorescence}\\
\hline
\end{tabular}
}
\caption{Error analysis of cross-task prompting. Four examples represent the major error characteristics (discussed in Section \ref{error-analysis}). The outputs shown are generated by LLaMA-2 13B. The examples shown here are shortened, complete examples are provided in Table \ref{tab:error-analysis-complete} of Appendix.}
\vspace{-5mm}
\label{tab:error-analysis}
\end{center}
\end{table*}

\section{Error analysis}\label{error-analysis}

We observe four types of error occurring in cross-task prompting (c.f. Table \ref{tab:error-analysis}), as follows:

% \begin{enumerate}[leftmargin=*,noitemsep,topsep=0pt]
    % \item 
    \noindent \textbf{Label space replication.} In example \#1, the source example is from the \textit{Conll2003-POS} dataset, a POS tagging task. In contrast, the target task is to predict the financial sentiment analysis. Therefore, the generation should be either negative, positive or neutral; however, we observe that the output is a sequence of POS tags instead. Here, the LLM is replicating the label space of the source task and the definition of the target task is not able to guide it to the correct target label space.
    
    % \item 
    \noindent \textbf{Junk prediction.} In some cases, we observe that the output is neither from the label space of the source task nor from that of the target task. As in example \#2, the source task is to classify each token into one named entity category (NER), and the target task is to predict the financial sentiment. The output in this case is found to be junk --- it is a sequence of \textit{N}'s but \textit{N} is not a pre-defined named entity category, neither is it a label for the target task. Further, we observe that the correct label in this example is `neutral' -- the LLM could not follow the target task definition provided and gets confused between the correct prediction label of the target task and the token classification task where the output is expected to be a sequence of labels, and instead outputs a sequence with the initial letter of the correct label.
    
    % \item 
    \noindent \textbf{Copying effect:} We also notice the copying effect, which has been observed by \citet{zicl}, where the predicted label is the same as the label of the context example which is semantically very similar to the target input. In example \#3, the target input is semantically very similar to the context example from the source task provided in the prompt, and consequently, the LLM incorrectly outputs the same label as that of the context example.
    
    % \item 
    \noindent \textbf{Definition not followed:} We observe that for some instances the LLM does not adhere to the definition given in the form of task definition for the target task. In example \#4, the LLM is supposed to output one among the four options \textit{A}, \textit{B}, \textit{C}, \textit{D}, instead it outputs the text corresponding to the option \textit{D} -- though \textit{D} is the correct answer in this case, the LLM is not able to follow the definition properly.
% \end{enumerate}
\section{Related work}

In-context learning without gradient updates to the model was introduced by \citet{DBLP:journals/corr/abs-2005-14165} using the GPT-3 model. Multiple recent works sought robust ICL setup via different techniques: selecting the examples that are semantically most similar to the input~\citep{liu-etal-2022-makes}, choosing low perplexity examples~\citep{gonen2022demystifying}, training a two-stage retrieval system to select the best examples~\citep{rubin-etal-2022-learning}, etc. However, these works primarily aim to construct better in-task prompts where the examples and the input come from the same task. \citet{tanwar-etal-2023-multilingual} showed that cross-lingual ICL can be elicited with proper alignment between the source and target language examples.
\citet{T5} introduced the {\em T5 transformer} model which has been trained to perform different text-processing tasks. \citet{zhang-etal-2022-task} proposed a task prefix guided multi-task pre-training framework to solve this problem. More recently, a new prompting method called {\em Meta-CoT} has been proposed by \citet{zou2023metacot} which generalizes chain-of-thought prompting to include multi-task examples in the prompt and it has shown improvement in a number of tasks.

% Drawing intuition from these works, we have explored the cross-task setup in ICL in this paper.

\label{sec:bibtex}
\section{Conclusion}

In this paper, we addressed LLMs' adaptability to novel tasks. Exploiting the inherent ICL capabilities of these models, we established that LLMs can substantially enhance their performance in novel task scenarios, even when direct examples are lacking. This encouraging outcome unveils fresh possibilities for the practical integration of LLMs across a broader spectrum of applications.
Furthermore, our study demonstrated the significance of generating pseudo-labels using cross-task prompting, presenting a potential solution for situations where annotated data is scarce. The observed correlation between the impact of cross-task examples and the similarity in model activations between source and target input tokens offers valuable insights into the underlying mechanisms of this phenomenon.
\section*{Limitations}

Despite presenting a potential future direction towards training-free task generalization using LLMs, this study has some important limitations. It is evident that the similarity between the source and the target tasks plays an important role in the performance. Hence, in real-world scenarios where the task novelty is extreme, such a method may fail to provide suitable performance. This is directly related to the fact that ours is the first study in this direction, and we have primarily focused on the empirical viability. A deeper understanding of generalizable task information captured inside the LLM circuits would help to come up with sophisticated solutions. Task novelty in our discussion does not presuppose access to novel knowledge. Hence, one can not mitigate the gap if a novel task requires the model to access newer information not present in the pretraining data or the in-context examples.
\bibliography{anthology,custom}
\bibliographystyle{acl_natbib}

\appendix

\section{Dataset details}
\label{sec:appendix-dataset}
\subsection{Source datasets}
We have used the following datasets as source datasets: 
\begin{itemize}
    \item[] \textbf{ARC-Easy:} ARC-Easy \cite{Clark2018ThinkYH} is a part of the ARC (AI2 Reasoning Challenge) dataset consisting of easy natural science questions targeted for students of 3rd to 9th grade. The questions are of multiple-choice format, where one of the four given options is correct. The training set consists of 2251 questions that we use for selecting our source examples. 
    
    \item[] \textbf{AG-news:} AG-news \cite{Zhang2015CharacterlevelCN} is a text classification dataset containing news articles categorized into four classes - \textit{sports}, \textit{business}, \textit{technology}, and \textit{world}. It has a training set size of 120\textit{K}, from which we have randomly sampled 10\textit{K} news articles to construct the source dataset for our experiments.
    
    \item[] \textbf{BoolQ:} BoolQ \cite{clark-etal-2019-boolq} is a reading comprehension dataset consisting of yes/no questions asked from a passage given for each question. The questions are mostly non-factoid, and considerable inference ability is required to answer them based on the passages provided. In our usage, each question is labeled as either \textit{true} or \textit{false}. The training set consists of 9427 labeled question-passage pairs from which we select the source examples.
    
    \item[] \textbf{Commonsense-QA:} Commonsense-QA \cite{talmor-etal-2019-commonsenseqa} is a commonsense question-answering dataset that consists of multiple-choice questions where one of the five options provided is correct. To answer the questions, logical reasoning abilities and in some cases prior knowledge are required. The training set consists of 9740 labeled questions from which source examples are chosen by us.

    \item[] \textbf{Conll2003-POS:} Conll2003-POS \cite{tjong-kim-sang-de-meulder-2003-introduction} is a collection of the data which is a part of the CoNLL-2003 shared task. In this dataset, each sentence is labeled as a sequence of part-of-speech (POS) tags (each token is assigned a POS tag). We construct our source dataset by sampling 10\textit{K} sentences from the 14,041 sentences in the training set.

    \item[] \textbf{Conll2003-NER:} Conll2003-NER \cite{tjong-kim-sang-de-meulder-2003-introduction} is also a part of the CoNLL-2003 shared task. Here, each sentence is labeled as a sequence of named entity tags. The task in this case is to perform named-entity recognition (NER). IOB tagging scheme \cite{iob} is used for assigning the tags. Four types of entities are assumed to be present in the data -- persons (PER), organizations (ORG), locations (LOC) and miscellaneous names (MISC). The training set consists of 14,041 sentences of which 10\textit{K} are sampled by us as our source data.

    \item[] \textbf{MNLI:} Multi-Genre Natural Language Inference (MNLI) \cite{N18-1101} corpus is one of the largest available resources for natural language inference. The corpus consists of 393\textit{K} labeled examples, where each example consists of a pair of sentences and the label is one among \textit{neutral}, \textit{contradiction} or \textit{entailment} based on the relationship between their meanings. Our source dataset is constructed by sampling 10\textit{K} examples from the 393\textit{K} examples available.

    \item[] \textbf{QQP:} Quora Question Pairs (QQP) \cite{DBLP:journals/corr/abs-1907-01041} dataset is curated for the task of natural language understanding. This dataset consists of question pairs collected from the popular question-answering website {\em Quora}, and the task is to determine if the questions are duplicates of each other. The training set consists of 364\textit{K} question pairs, each labeled as \textit{duplicate} or \textit{not duplicate}. We sample 10\textit{K} labeled question pairs from the training set for our source data.

    \item[] \textbf{RACE:} RACE \cite{lai-etal-2017-race} is a reading comprehension dataset consisting of passages along with questions asked from them. The passages and questions are collected from English exams of school students aged between 12 to 18. The questions are of multiple-choice format where one of the four options is correct. Our source dataset is gathered by picking 10\textit{K} passage-question pairs from 87.9\textit{K} such pairs available in the training set.
    
    \item[] \textbf{SST2:} SST2 \cite{socher-etal-2013-recursive} dataset is a part of the Stanford Sentiment Treebank corpus. It contains movie review snippets collected from the \textit{Rotten Tomatoes} website. Each review is labeled as \textit{positive} or \textit{negative}. The 10\textit{K} reviews for our source dataset are sampled from the training set consisting of 67.3\textit{K} labeled reviews.
    
\end{itemize}

\subsection{Target datasets}
We have used the following datasets as target datasets:

\begin{itemize}
    \item[] \textbf{ARC-Challenge:} ARC-Challenge \cite{Clark2018ThinkYH} is also a part of the ARC (AI2 Reasoning Challenge) dataset and it consists of hard natural science questions targeted for students of 3rd to 9th grade. Those questions that were answered incorrectly by both a retrieval-based algorithm and a word co-occurrence method are considered to be \lq hard\rq enough for inclusion in this dataset. The questions are of multiple-choice format, where one of the four options is correct. The test set consists of 1172 questions out of which 500 were randomly selected for our target dataset.

    \item[] \textbf{Social-i-QA:} Social-i-QA \cite{DBLP:journals/corr/abs-1904-09728} is a commonsense reasoning dataset focusing on social situations. This dataset contains examples consisting of a social situation or action given as context and a multiple-choice question asked based on the context aimed at testing emotional and social intelligence. Each question has three options, out of which one is correct.
    We select 500 examples for our target data from the 1954 examples available in the validation set.

    \item[] \textbf{SciQ:} SciQ \cite{SciQA2023} (or, SciQA) is a scientific question-answering dataset consisting of questions from different areas of Physics, Chemistry and Biology. The questions are of multiple-choice format, where one of the four options is correct. The test set consists of 1000 questions, of which 500 are sampled to prepare our target dataset.

    \item[] \textbf{MedMCQA:} MedMCQA \cite{pmlr-v174-pal22a} is a multiple-choice question-answering dataset consisting of questions from post-graduate medical entrance exams in India. Each question has four options of which one is correct. Our target dataset is constructed by selecting 500 questions from the 4183 questions in the validation set. 
    
    \item[] \textbf{Financial-Phrasebank:} Financial-Phrasebank \cite{Malo2014GoodDO} is a financial sentiment analysis dataset containing sentences mined from a corpus of English news on all listed companies in OMX Helsinki. Each sentence is labeled as one of the three categories -- \textit{positive}, \textit{negative}, or \textit{neutral}, based on the influence the news snippet may have on the stock price. The training set consists of 2264 labeled sentences, from which 500 are sampled for our target dataset.

\end{itemize}

In each case, for preparing our target dataset we have selected 500 examples. The selection, though random, is done in such a way that our target datasets are balanced, i.e. the number of examples with each of the different possible labels is almost equal. 

\section{Semantic context creation}
\label{sec:semantic-context-creation}
{\color{black}One might assume that taking internal embeddings of the models, instead of an external model like Sentence-BERT, for extracting semantically similar examples might be better suited. However, we found that doing so for LLaMA-2 7B is computationally more expensive, with no significant gain in performance, as reported in Table~\ref{tab:llama-sim}.

%One might assume taking internal embedding of the models instead of an external model like Sentence-BERT, for extracting semantically similar examples might be better suited, but we found that doing so for LLaMA-2 7B was specially and computationally expensive, with no significant gain in performance Table \ref{tab:llama-sim}.

We also experimented with Sentence-BERT and E5 \cite{e5} to test which external model produces superior semantically similar context $C$ for the LLMs to use. Table~\ref{tab:e5} shows the results for LLaMA-2 7B and GPT3.5 performance using $C$ from the E5 model. We proceeded with Sentence-BERT instead of E5 as our external model, as the former has already been used for a similar role in prior works \cite{liu-etal-2022-makes,tanwar-etal-2023-multilingual}, and its performance is also comparable to that of E5.
% Apart from Sentence-BERT, we also experimented by considering the final hidden layer representation of source and target inputs from LLaMA-2 7B and E5 \cite{e5} for selecting the similar source examples -- the corresponding results obtained are reported in Tables \ref{tab:llama-sim} and \ref{tab:e5} of Appendix, respectively. We observe almost similar accuracy values and consistent trend in the results across all three models, viz. Sentence-BERT, LLaMA-2 7B and E5, when utilizing their embeddings for similar source example selection.
}

%%%%%%%%%%%%%%%%%%%%%%%%%%%%%%%% LLaMA-2 7B similarity experiment %%%%%%%%%%%%%%%%%%%%%%%%%%%%%%%%%%%%%%%%%
\begin{table*}[!t]
\small
\centering
\adjustbox{max width=1\linewidth}{
{\color{black} \begin{tabular}{lccccccccccc}
\hline
 \backslashbox{TAR}{SRC}& Zero-shot & ARC-Easy & AG-news & BoolQ & Com-QA & C-POS & C-NER & MNLI & QQP  & RACE  & SST2 \\ \hline
\multicolumn{12}{c}{\cellcolor[HTML]{C0C0C0}LLaMA-2 7B}                                                                                \\ \hline

ARC-Challenge  & 4.6 & \textbf{44}  & 32.2  & 36.6  & 43.2  & 36.8  & 35.2  & 34.6  & 33.8  & 37.4  & 33.2\\ 
Financial-Phrasebank  & 34.1 & 51.1  & 58.28  & 39.9  & 11.2  & 0.2  & 0  & \textbf{65.6}  & 42.31  & 40.3  & 60.4 \\
MedMCQA & 4.2 & 30.2  & 25.8  & 27  & \textbf{32.4}  & 24.6  & 25.2  & 24.8  & 25.6  & 30.6  & 25.2\\ 
SciQ  & 8.0 & 60.2  & 44.8  & 46.6  & \textbf{66.8}  & 33.6  & 28  & 47.8  & 25.6  & 61.8  & 42 \\ 
Social-i-QA  & 41.1 & 41.7  & 35.5  & 37.3  & \textbf{47.3}  & 39.3  & 37.9  & 37.9  & 36.3  & \textbf{47.3}  & 39.7 \\ \hline
%%%

\end{tabular}}
}
\captionsetup{labelfont={color=black},font={color=black}}
\caption{Accuracy for cross-task setup using one source example which is most similar to the target input based on their final hidden layer representation from LLaMA-2 7B. (Abbreviations: Com-QA \(\rightarrow\) Commonsense-QA, C-POS \(\rightarrow\) Conll2003-POS, C-NER \(\rightarrow\) Conll2003-NER).}
\label{tab:llama-sim}
\vspace{-1mm}
\end{table*}

%%%%%%%%%%%%%%%%%%%%%%%%%%%%%%%% E5 experiment %%%%%%%%%%%%%%%%%%%%%%%%%%%%%%%%%%%%%%%%%
\begin{table*}[!t]
\small
\centering
\adjustbox{max width=1\linewidth}{
{\color{black} \begin{tabular}{lccccccccccc}
\hline
 \backslashbox{TAR}{SRC}& Zero-shot & ARC-Easy & AG-news & BoolQ & Com-QA & C-POS & C-NER & MNLI & QQP  & RACE  & SST2 \\ \hline
\multicolumn{12}{c}{\cellcolor[HTML]{C0C0C0}LLaMA-2 7B}                                                                                \\ \hline
ARC-Challenge & 4.6 & 44.80 & 34.20 & 36.00 & 44.00 & 32.00 & 31.80 & 35.80 & 35.20 & \textbf{48.80} & 33.60 \\ 
Financial-Phrasebank & 34.1 & 43.91 & 60.08 & 38.92 & 38.32 & 4.39 & 0.00 & \textbf{67.86} & 56.09 & 50.30 & 46.71 \\ 
MedMCQA & 4.2 & 30.40 & 24.40 & 28.20 & \textbf{32.80} & 24.20 & 24.60 & 23.60 & 25.60 & 29.20 & 23.40 \\ 
SciQ & 8.0 & 62.40 & 38.80 & 46.00 & \textbf{68.40} & 31.80 & 32.40 & 36.60 & 27.60 & 65.40 & 37.20 \\ 
Social-i-QA & 41.1 & 42.51 & 37.33 & 37.92 & 46.91 & 37.72 & 40.52 & 41.72 & 39.52 & \textbf{47.90} & 41.52 \\ \hline
%Average  & 18.4 & 44.80 & 38.96 & 37.41 & 46.09 & 26.02 & 25.86 & 41.12 & 36.80 & \textbf{48.32} & 36.49 \\ \hline

%%%%

\multicolumn{12}{c}{\cellcolor[HTML]{C0C0C0}GPT3.5}                                                                               \\ \hline

ARC-Challenge & 74.6 & 77.00 & 74.80 & 75.80 & 76.60 & 70.20 & 69.80 & 72.40 & 74.40 & \textbf{77.40} & 73.00 \\ 
Financial-Phrasebank & 57.5 & 78.44 & \textbf{82.63} & 75.25 & 77.45 & 63.67 & 70.46 & 67.66 & 74.25 & 76.65 & 63.67 \\ 
MedMCQA  & 49.6  & 47.00 & 49.60 & 49.20 & \textbf{49.80} & 48.40 & 46.20 & 47.20 & 46.60 & 49.60 & 46.80 \\ 
SciQ & 91.2 & 91.20 & 89.00 & 91.00 & 91.00 & 85.80 & 84.60 & 89.20 & 89.20 & \textbf{93.40} & 90.60 \\ 
Social-i-QA & 76.0 & 77.05 & 74.65 & 74.85 & 77.05 & 73.85 & 73.05 & 74.05 & 73.25 & \textbf{77.25} & 71.86 \\ \hline
%Average & 69.8 & 74.14 & 74.14 & 73.22 & 74.38 & 68.38 & 68.82 & 70.10 & 71.54 & 74.86 & 69.19 \\ \hline
%%%

\end{tabular}}
}
\captionsetup{labelfont={color=black},font={color=black}}
\caption{Accuracy for cross-task setup using one source example which is most similar to the target input based on their embeddings generated by the E5 model. (Abbreviations: Com-QA \(\rightarrow\) Commonsense-QA, C-POS \(\rightarrow\) Conll2003-POS, C-NER \(\rightarrow\) Conll2003-NER).}
\label{tab:e5}
\vspace{-1mm}
\end{table*}

%%%%%%%%%%%%%%%%%%%%%%%%%%%%%%%%% INSTRUCTION-TUNED MODEL %%%%%%%%%%%%%%%%%%%%%%%%%%%%
\section{Cross-task prompting in instruction-tuned models}
\label{sec:instruction-tuned}
The performance gain with cross-task prompting for instruction-tuned models is lesser compared to that for the corresponding base model, due to the improvement in zero-shot performance of instruction-tuned models. For instance, Table \ref{tab:ins-tuned} shows the performance for LLaMA-2 7B Chat model. We observe that it’s zero-shot performance is much better compared to the LLaMA-2 7B base model.

\begin{table*}[!t]
\small
\centering
\adjustbox{max width=1\linewidth}{
{\color{black} \begin{tabular}{lccccccccccc}
\hline
 \backslashbox{TAR}{SRC}& Zero-shot & ARC-Easy & AG-news & BoolQ & Com-QA & C-POS & C-NER & MNLI & QQP  & RACE  & SST2 \\ \hline
\multicolumn{12}{c}{\cellcolor[HTML]{C0C0C0}LLaMA-2 7B Chat}                                                                                \\ \hline
ARC-Challenge & 42.00 & 51.00 & 39.20 & 45.00 & \textbf{51.40} & 37.00 & 30.20 & 37.00 & 37.20 & \textbf{51.40} & 39.00 \\ 
Financial-Phrasebank & 79.64 & 76.05 & \textbf{83.63} & 77.84 & 68.66 & 43.91 & 12.18 & 77.25 & 80.84 & 78.84 & 83.43 \\ 
MedMCQA & 30.60 & 31.00 & 30.20 & 31.60 & \textbf{34.60} & 30.00 & 26.80 & 30.20 & 27.40 & 31.00 & 30.20 \\ 
SciQ & 65.60 & 66.60 & 59.00 & 67.00 & \textbf{71.80} & 58.20 & 45.00 & 53.80 & 60.80 & 69.20 & 59.60 \\ 
Social-i-QA & 52.10	& 48.10	& 47.70	& 47.70	& 49.50	& 47.90	& 36.33	& 42.32	& 41.92	& \textbf{53.69} & 45.91 \\ \hline
\end{tabular}}
}
\captionsetup{labelfont={color=black},font={color=black}}
\caption{Accuracy for cross-task setup, using one source example, with the LLaMA-2 7B Chat model. (Abbreviations: Com-QA \(\rightarrow\) Commonsense-QA, C-POS \(\rightarrow\) Conll2003-POS, C-NER \(\rightarrow\) Conll2003-NER).}
\label{tab:ins-tuned}
\vspace{-1mm}
\end{table*}
%%%%%%%%%%%%%%%%%%%%%%%%%%%%%%%%%%%%%%%%%%%% FORCE DECODING %%%%%%%%%%%%%%%%%%%%%%%%%%%%%%%%%%%%%%%%%%%%%%%
\section{Force decoding}
\label{sec:force-decoding}
{\color{black}%\noindent \textbf{Forced Decoding.}
Unlike greedy decoding, where the most probable token from the entire vocabulary of the model is the output, in force decoding the vocabulary is restricted to a set of tokens and the output is assigned out of these restricted tokens. Formally, for context $C$ and a target task $T$, the output in the case of force decoding is:

\begin{equation*}
\small
    \hat{y}_t = \underset{y \in L_T}{\mathrm{argmax}}\, p(y|C)
\end{equation*}

where, $L_T$ is the label space of target task $T$. The label space of all five target tasks is shown in Table \ref{tab:label-space}.

Table \ref{tab:force-decode} reports the performance of cross-task prompting for all source-target pairs using force decoding in LLaMA-2 7B and LLaMA-2 13B models. We noted that force decoding improves the performance of zero-shot prediction by a great margin. However, the same is not true for every source-target pair.
}

\begin{table*}[!t]
\small
\centering
\adjustbox{max width=1\linewidth}{
{\color{black}\begin{tabular}{lccccccccccc}
\hline
 \backslashbox{TAR}{SRC}& Zero-shot & ARC-Easy & AG-news & BoolQ & Com-QA & C-POS & C-NER & MNLI & QQP  & RACE  & SST2 \\ \hline
\multicolumn{12}{c}{\cellcolor[HTML]{C0C0C0}LLaMA-2 7B}                                                                                \\ \hline
ARC-Challenge 	& 38.80 	& 45.40 	& 33.60 	& 36.00 	& 45.00 	& 35.20 	& 32.80 	& 35.40 	& 35.20 	& \textbf{46.60} 	& 35.00 \\
Financial-Phrasebank 	& 36.33 	& 47.11 	& \textbf{63.27} 	& 60.28 	& 45.91 	& 50.50 	& 53.49 	& 62.87 	& 57.29 	& 44.91 	& 53.09 \\
MedMCQA 	& 26.00 	& 28.00 	& 26.60 	& 28.80 	& \textbf{32.60} 	& 28.20 	& 26.80 	& 27.00 	& 27.00 	& 29.60 	& 27.40 \\
SciQ 	& 59.60 	& 61.60 	& 37.00 	& 49.80 	& \textbf{65.60}	& 37.00 	& 31.60 	& 33.20 	& 27.20 	& 62.00 	& 32.40 \\
Social-i-QA 	& 47.11 	& 43.31 	& 41.32 	& 39.12 	& \textbf{48.90} 	& 40.92 	& 40.92 	& 43.11 	& 39.72 	& 48.10 	& 40.52 \\ \hline

\multicolumn{12}{c}{\cellcolor[HTML]{C0C0C0}LLaMA-2 13B}                                                                               \\ \hline
ARC-Challenge 	& 52.80 	& \textbf{58.80} 	& 54.60 	& 56.00 	& 56.80 	& 51.20 	& 50.20 	& 53.40 	& 54.00 	& 57.20 	& 52.00 \\
Financial-Phrasebank 	& 62.08 	& 73.25 	& 79.64 	& 70.06 	& 66.87 	& 66.27 	& 67.47 	& 75.65 	& \textbf{80.24} 	& 70.86 	& 77.64 \\
MedMCQA 	& 36.60 	& 38.20 	& 36.00 	& 37.60 	& \textbf{38.80} 	& 36.00 	& 32.20 	& 33.80 	& 31.20 	& 39.60 	& 33.00 \\
SciQ 	& 82.00 	& \textbf{83.60} 	& 81.40 	& 82.80 	& 82.20 	& 80.60 	&  79.00 	& 81.20 	& 79.80 	& 82.80 	& 77.80 \\ 
Social-i-QA 	& 57.68 	& 62.08 	& 58.08 	& 56.49 	& 62.28 	& 57.49 	& 57.49 	& 55.29 	& 56.49 	& \textbf{65.07} 	& 57.88 \\ \hline

\end{tabular}
}}
\captionsetup{labelfont={color=black},font={color=black}}
\caption{Accuracy for cross-task setup using one source example by force decoding, where the label from the label space to which the LLM assigns the highest probability is considered as the output. (Abbreviations: Com-QA \(\rightarrow\) Commonsense-QA, C-POS \(\rightarrow\) Conll2003-POS, C-NER \(\rightarrow\) Conll2003-NER).}
\label{tab:force-decode}
\vspace{-5mm}
\end{table*}

%%%%%%%%%%%%%%%%%%%%%%%%%%%% Label Space %%%%%%%%%%%%%%%%%%%%%%%%%%%%
\begin{table*}[!t]
\small
\begin{center}
\scalebox{1}{
{\color{black}\begin{tabular}{ |m{3cm}|m{5cm}|}
 \hline
 {\bf Target task}&{\bf Label space} \\
  \hline

ARC-Challenge & \{A, B, C, D\}\\
\hline

Financial-Phrasebank & \{positive, neutral, negative\}\\
\hline

MedMCQA & \{A, B, C, D\}\\
\hline

SciQ & \{A, B, C, D\}
\\
\hline

Social-i-QA & \{A, B, C\}
\\
\hline
\end{tabular}}
}
\captionsetup{labelfont={color=black},font={color=black}}
\caption{Label space of target tasks}
\vspace{-2mm}
\label{tab:label-space}
\end{center}
\end{table*}

%%%%%%%%%%%%%%%%%%%%%%%%%%%%%%%%%%%%%%% RANDOM LABEL %%%%%%%%%%%%%%%%%%%%%%%%%%%%%%%%%%%%%%%%
\section{Random labeling}
\label{sec:random-label-icl}
\textcolor{black}{Recently, \citet{symbol-tune} proposed using a random label space for pseudo-labeling examples of (target) task and utilise these to generate the context $C$. We experimented with this setup as it is a better alternative to zero-shot prompting, but our results (Table \ref{tab:random-label-icl}) showed that the model output is random with such $C$.}

%%%%%%%%%%%%%%%%%%%%%%%%%%%%%%%% random label space %%%%%%%%%%%%%%%%%%%%%%%%%%%%%%%%%%%%%%%%%
\begin{table*}[!t]
\small
\centering
\adjustbox{max width=1\linewidth}{
{\color{black} \begin{tabular}{lccccccccccc}
\hline
 \backslashbox{Shots}{TRC}& ARC-Challenge & Financial-Phrasebank & MedMCQA & SciQ & Social-i-QA \\ \hline
\multicolumn{6}{c}{\cellcolor[HTML]{C0C0C0}LLaMA-2 7B}                                                                                \\ \hline
1 &25.07&33.33&25.00&25.00&33.27 \\
8 &25.00&34.13&24.80&25.40&32.73\\ \hline

\multicolumn{6}{c}{\cellcolor[HTML]{C0C0C0}LLaMA-2 13B}                                                                                \\ \hline
1 &25.13&31.00&25.00&25.07&33.33 \\
8 &24.40&33.33&26.00&25.20&33.53\\ \hline

\multicolumn{6}{c}{\cellcolor[HTML]{C0C0C0}GPT3.5}                                                                                \\ \hline
1 & 24.00 & 49.50 & 22.60 & 26.60 & 32.58 \\
8 & 24.00 & 56.89 & 24.40 & 37.00 & 33.13\\ \hline

\end{tabular}}
}
\captionsetup{labelfont={color=black},font={color=black}}
\caption{Accuracy for random label space labelling. We experimented with random labeled context $C$ of size one and eight.}
\label{tab:random-label-icl}
\vspace{-1mm}
\end{table*}

\begin{table*}[!t]
\small
\centering
\adjustbox{max width=1\linewidth}{
\begin{tabular}{lcc}
\hline
                                         & With source definitions                  & W/O source definitions   \\ \hline
\multicolumn{3}{c}{\cellcolor[HTML]{C0C0C0}LLaMA-2 7B}                                                               \\ \hline
ARC-Challenge                            &        $37.0$                &     $33.0$                           \\
Financial-Phrasebank                     &        $38.8$          &     $27.0$           \\
MedMCQA                                  &         $27.3$          &     $17.06$           \\
SciQ                                     &        $45.7$          &     $48.1$            \\ 
Social-i-QA                              &        $41.8$          &     $42.0$            \\ \hline
Average                                  &       $38.1$          &     $33.8$            \\ \hline
\multicolumn{3}{c}{\cellcolor[HTML]{C0C0C0}LLaMA-2 13B}                                                               \\ \hline
ARC-Challenge                            &        $53.2$                &     $52.42$              \\
Financial-Phrasebank                     &        $58.9$          &     $48.0$            \\
MedMCQA                                  &         $34.2$          &     $24.36$           \\
SciQ                                     &        $78.2$          &     $75.3$            \\ 
Social-i-QA                              &        $57.5$          &     $58.1$           \\ \hline
Average                                  &       $56.4$          &     $51.6$       \\ \hline
\end{tabular}
}
\caption{Accuracy With and Without (W/O) source definition. We note a drop in accuracy when source definitions are removed.}
\label{tab:no_source_inst}
\vspace{-5mm}
\end{table*}

\section{Prompt details}
\label{sec:appendix}

We show a few examples of cross-task prompts in Table \ref{tab:cross-task-prompts} and in-task combined with cross-task prompts in Table \ref{tab:icl-with-cross-task}. Additionally, task definitions of source and target tasks are provided in Table \ref{tab:source-task-definitions} and Table \ref{tab:target-task-definitions} respectively.

%%%%%%%%%%%%% CROSS - TASK %%%%%%%%%%%%%

\begin{table*}[!t]
\small
\begin{center}
\scalebox{0.9}{
\begin{tabular}{ |m{1.4cm}|m{2cm}|m{12.4cm}| m{0.8cm}|}
 \hline
 {\bf Target task}&{\bf Source task}&{\bf Prompt}&{\bf Output} \\ \hline
Financial-Phrasebank & QQP & Definition: Given two question pairs do text classification based on whether they are duplicates or not. The questions are mined from the popular online discussion forum Quora. As duplicate questions might be present on Quora, the task is to label two identical questions as "duplicate" if they ask the same query else label the pair as "not duplicate".

Question 1: My yearly income went from \$0 to \$55,000. By how many times did it increase? 

Question 2:What can I do with approximately 10.000 Baht per month to increase my income? 

Label: not duplicate

Definition: Given a sentence mined from a financial news article, you are to determine the sentiment polarity of the sentence. The task deals with financial sentiment analysis. Based on the sentiment conveyed by the sentence, label the sentence as "negative", "positive" or "neutral"

Sentence: Net income from life insurance doubled to EUR 6.8 mn from EUR 3.2 mn , and net income from non-life insurance rose to EUR 5.2 mn from EUR 1.5 mn in the corresponding period in 2009 . 

Label: & \textcolor{red}{positive} \\
\hline

SciQ & Commonsense-QA & Definition: The following task relates to commonsense reasoning. It consists of a question that can be easily solved using logical abilities and reasoning, a set of five options  "A.", "B.", "C.", "D." and "E." are also provided along with the question, one of these options answers the question logically. Use your reasoning ability to select the most appropriate answer from the provided choices "A.", "B.", "C.", "D." and "E." and assign these choices (i.e  "A.", "B.", "C.", "D." and "E.") as the label

Question:In what substance do clouds float? 

A. sky 

B. top of mountain 

C. air 

D. ground level 

E. outer space 

Answer: C

Definition: Given a question from a scientific exam about Physics, Chemistry, and Biology, among others. The question is in multiple choice format with four answer options "A.", "B.", "C." and "D.". Using your knowledge about the scientific fields answer the question and provide the label "A", "B", "C" and "D" as answer

Question: What term means the amount of water vapor in the air?

A. pressure

B. humidity

C. temperature

D. ambient 

Answer: & \textcolor{red}{B} \\
\hline

Social-i-QA & RACE & Definition: Given a reading comprehension type question-answering from an english exam for school students. You are given a context and multiple choice question containing four options "A.", "B.", "C." and "D.". The question is answerable from the comprehension. Based on the question, the option and the context select the most appropriate answer from the provided choices "A.", "B.", "C." and "D.".

Context: Mike is a factory worker. He is often very tired after a day's work. His wife, Jenny, has no job, so she stays at home to cook the meals. Every day he can have his dinner when he gets home from his factory.
One day, Mike came home very late because he was very busy in the factory. He was very hungry when he got home.
He was not happy when he found his dinner was not ready. He was very angry with his wife. He shouted at her, "I'm going out to eat in a restaurant." "Wait for five minutes," said his wife. "Why? Do you think that dinner will be ready in five minutes?" asked Mike.
"Of course not," she answered. "But I can be ready to go with you in five minutes." 

Question: Mike works in \_  . 

A. a factory 

B. an office 

C. a school 

D. a hospital 

Answer: A

Definition: Given an action as the context and a related question, you are to answer the question based on the context using your social intelligence. The question is of multiple choice form with three options "A", "B" and "C". Select the most appropriate answer from the provided choices "A", "B" and "C".

Context: Jesse is patient and hardworking and hungry in the morning. 

Question: How would you describe Jesse? 

A. ill prepared 

B. Exhausted and starved 

C. thoughtful 

Answer: & \textcolor{red}{B} \\
\hline
\end{tabular}
}
\caption{Examples of prompts for cross-task prompting technique. The number of source examples in each case is one. The outputs shown are generated by LLaMA-2 13B.}
\vspace{-2mm}
\label{tab:cross-task-prompts}
\end{center}
\end{table*}

%%%%%%%%%%%%%%%%%%% ICL + Cross-Task %%%%%%%%%%%%%%%%

\begin{table*}[!t]
\small
\begin{center}
\scalebox{0.9}{
\begin{tabular}{ |m{1.4cm}|m{1.4cm}|m{13cm}| m{0.8cm}|}
 \hline
 {\bf Target task}&{\bf Source task}&{\bf Prompt}&{\bf Output} \\
  \hline
Financial-Phrasebank & SST2 & Definition: Given a movie review do text classification, based on the sentiment conveyed by the review label it as "positive" or "negative"

Sentence: results is the best performance from either in years  

Label: positive

Definition: Given a sentence mined from a financial news article, you are to determine the sentiment polarity of the sentence. The task deals with financial sentiment analysis. Based on the sentiment conveyed by the sentence, label the sentence as "negative", "positive" or "neutral"

Sentence: For the last quarter of 2010 , Componenta 's net sales doubled to EUR131m from EUR76m for the same period a year earlier , while it moved to a zero pre-tax profit from a pre-tax loss of EUR7m .

Label: positive 
Sentence: Both operating profit and net sales for the six-month period increased , respectively from EUR18 .1 m and EUR127 .6 m , as compared to the corresponding period in 2006 . 

Label: & \textcolor{red}{positive} \\
\hline

MedMCQA & RACE & Definition: Given a reading comprehension type question-answering from an english exam for school students. You are given a context and multiple choice question containing four options "A.", "B.", "C." and "D.". The question is answerable from the comprehension. Based on the question, the option and the context select the most appropriate answer from the provided choices "A.", "B.", "C." and "D.".

Context: Plants need green leaves to make food. A plant needs sunlight and carbon dioxide  from the air for making food and it also needs water and salts from the soil to make food too. There are certain cells   in the leaves which change carbon dioxide and water into sugar. To do this the cells needs energy, which they get from the sunlight.
Green leaves make food for the whole plant. A red leaf can make food too because under the red color1ing of the leaf there are food----making cells. There are no leaves which are completely yellow, for they can't make food.
The plant makes sugar for its food. In sunlight green leaves make a lot of sugar. The veins   can't carry all this sugar away, so the leaves change the sugar into starch, which is kept and so stored in the leaves. At night, the starch changes back to sugar. It is then carried away from the leaves. Some of the sugar is used as food by the plant while the rest is stored as starch. In some plants, food is stored in the roots, in others it is stored in the stem   and in leaves, fruits and seeds. 

Question:Sugar is made for its food by - 

A. sunlight 

B. veins 

C. stems 

D. green leaves 

Answer: D

Definition: Given a multiple choice question containing four options "A.", "B.", "C." and "D." from a medical entrance exam. The question is related to a sub-field of medical science like Microbiology, Radiology, Ophthalmology, Surgery, Human anatomy, etc. Based on the question, the option and your knowledge of the medical field select the most appropriate answer from the provided choices "A.", "B.", "C." and "D.".

Question: Growth hormone has its effect on growth through?

A. Directly 

B. IG1-1 

C. Tyroxine 

D. Intranuclear receptors 

Answer: B 

Question:Which of the following has intracellular receptor - 

A. Glucagon 

B. Insulin 

C. Epinephrine 

D. Thyroxine 

Answer: & \textcolor{red}{D} \\
\hline

SciQ & ARC-Easy & Definition: Given a question answering task from the 3rd to 9th-grade science exam. The question contains four options "A.", "B.", "C." and "D." Select the most appropriate choice that answers the question

Question: From which part of the plant does a bee get food?

A. flower

B. seed

C. stem

D. root 

Answer: A

Definition: Given a question from a scientific exam about Physics, Chemistry, and Biology, among others. The question is in multiple choice format with four answer options "A.", "B.", "C." and "D.". Using your knowledge about the scientific fields answer the question and provide the label "A", "B", "C" and "D" as answer

Question: What type of organism is commonly used in preparation of foods such as cheese and yogurt?

A. Viruses 

B. Protozoa 

C. Gymnosperms 

D. Mesophilic organisms 

Answer: D 

Question: A bee will sometimes do a dance to tell other bees in the hive where to find what?

A. water

B. food

C. honey

D. enemies 

Answer: & \textcolor{red}{B} \\
\hline
\end{tabular}
}
\caption{Examples of prompts for in-task prompting combined with cross-task prompting technique. The outputs shown are generated by LLaMA-2 13B.}
\vspace{-2mm}
\label{tab:icl-with-cross-task}
\end{center}
\end{table*}

%%%%%%%%%%%%%%%%%%%%%%%%%%%%%%%%%%%%%%%%%%%%%%%%%%%%%%%%%%%%%%% Source Task Definitions %%%%%%%%%%%%%%%%%%%%%%%%%%%%%%%%%%%%%%%%%%%%%%%%%%%%%%%%%%%%%%%

\begin{table*}[!t]
\small
\begin{center}
\scalebox{1}{
\begin{tabular}{ |m{2.4cm}|m{13.25cm}|}
 \hline
 {\bf Source task}&{\bf Task definition} \\
  \hline

AG-news & Given a sentence do text classification, the sentence is a clipping from a news article that may be either related to sports, business, technology, or world news. You are to recognize the category of the sentence and label them as "sports", "business", "technology" or "world" news 
 \\
\hline

ARC-Easy & Given a question answering task from the 3rd to 9th-grade science exam. The question contains four options "A.", "B.", "C." and "D." Select the most appropriate choice that answers the question
\\
\hline

BoolQ & Given a context and a question do binary true and false type text classification. You are given a passage as context and a question related to the passage that can be answered as "True" or "False". Based on the context, question and your reasoning ability answer in a "True" and "False".
\\
\hline

Commonsense-QA &The following task relates to commonsense reasoning. It consists of a question that can be easily solved using logical abilities and reasoning, a set of five options  "A.", "B.", "C.", "D." and "E." are also provided along with the question, one of these options answers the question logically. Use your reasoning ability to select the most appropriate answer from the provided choices "A.", "B.", "C.", "D." and "E." and assign these choices (i.e  "A.", "B.", "C.", "D." and "E.") as the label 

\\
\hline

Conll2003-NER &Given a sentence do token classification on it seek to locate and classify named entities mentioned in the sentence provided. The pre-defined named entity categories along with there labeles are Person (PER), Location (LOC), Organization (ORG) and Miscellaneous (MIS). If the token is not an entity mark it as None. As the entity is more than two tokens long use the prefix B with the named entity token to represent the beginning and  use the prefix I till the entity ends.
\\
\hline

Conll2003-POS &Given a sentence do token classification by doing Part-of-speech (POS) tagging, which is a process in natural language processing (NLP) where each word in a text is labeled with its corresponding part of speech. This can include nouns, verbs, adjectives, and other grammatical categories.
\\
\hline

MNLI &Given Sentence 1 which is a premise and Sentence 2 which is a hypothesis do natural language inference on the pair. In natural language inference we mark whether the premise and hypothesis are "neutral", "contradiction" or "entailment". The pair are said to be "entailed" if the premise justifies/supports the hypothesis, if the pair contradict each other we label them as "contradiction" and label them "neutral" in all other cases
\\
\hline

QQP &Given two question pairs do text classification based on whether they are duplicates or not. The questions are mined from the popular online discussion forum Quora. As duplicate quetion might be present on Quora, the task is to label two identical questions as "duplicate" if they ask the same query else label the pair as "not duplicate".
\\
\hline

RACE &Given a reading comprehension type question-answering from an english exam for school students. You are given a context and multiple choice question containing four options "A.", "B.", "C." and "D.". The question is answerable from the comprehension. Based on the question, the option and the context select the most appropriate answer from the provided choices "A.", "B.", "C." and "D.".
\\
\hline

SST2 &Given a movie review do text classification, based on the sentiment conveyed by the review label it as "positive" or "negative"
\\
\hline

\end{tabular}
}
\caption{Task definitions of source tasks}
\vspace{-2mm}
\label{tab:source-task-definitions}
\end{center}
\end{table*}

%%%%%%%%%%%%%%%%%%%%%%%%%%%%%%%%%%%%%%%%%%%%%%%%%%%%%%%%%%%%%%% Target Task Definitions %%%%%%%%%%%%%%%%%%%%%%%%%%%%%%%%%%%%%%%%%%%%%%%%%%%%%%%%%%%%%%%
\begin{table*}[!t]
\small
\begin{center}
\scalebox{1}{
\begin{tabular}{ |m{2.4cm}|m{13.25cm}|}
 \hline
 {\bf Target task}&{\bf Task definition} \\
  \hline

ARC-Challenge & Given a question answering task from the 3rd to 9th-grade science exam. The question contains four options "A.", "B.", "C." and "D." Select the most appropriate choice that answers the question\\
\hline

Financial-Phrasebank & Given a sentence mined from a financial news article, you are to determine the sentiment polarity of the sentence. The task deals with financial sentiment analysis. Based on the sentiment conveyed by the sentence, label the sentence as "negative", "positive" or "neutral"\\
\hline

MedMCQA & Given a multiple choice question containing four options "A.", "B.", "C." and "D." from a medical entrance exam. The question is related to a sub-field of medical science like Microbiology, Radiology, Ophthalmology, Surgery, Human anatomy, etc. Based on the question, the option and your knowledge of the medical field select the most appropriate answer from the provided choices "A.", "B.", "C." and "D.".\\
\hline

SciQ & Given a question from a scientific exam about Physics, Chemistry, and Biology, among others. The question is in multiple choice format with four answer options "A.", "B.", "C." and "D.". Using your knowledge about the scientific fields answer the question and provide the label "A", "B", "C" and "D" as answer
\\
\hline

Social-i-QA & Given an action as the context and a related question, you are to answer the question based on the context using your social intelligence. The question is of multiple choice form with three options "A", "B" and "C". Select the most appropriate answer from the provided choices "A", "B" and "C".
\\
\hline
\end{tabular}
}
\caption{Task definitions of target tasks}
\vspace{-2mm}
\label{tab:target-task-definitions}
\end{center}
\end{table*}

%%%%%%%%%%%%% Error analysis %%%%%%%%%%%%%
\section{Error analysis}
Table~\ref{tab:error-analysis-complete} contains detailed examples for erroneous predictions in cross-task prompting setup.

\begin{table*}[!t]
\small
\begin{center}
\scalebox{0.9}{
\begin{tabular}{ |m{0.1cm}|m{13cm}| m{2.5cm}|}
 \hline
 \#&{\bf Prompt}&{\bf Output} \\
  \hline
1.&Definition: Given a sentence do token classification by doing Part-of-speech (POS) tagging, which is a process in natural language processing (NLP) where each word in a text is labeled with its corresponding part of speech. This can include nouns, verbs, adjectives, and other grammatical categories.

Sentence: Net profit 6.08 vs 3.98 

Label: JJ NN CD NNP CD

Definition: Given a sentence mined from a financial news article, you are to determine the sentiment polarity of the sentence. The task deals with financial sentiment analysis. Based on the sentiment conveyed by the sentence, label the sentence as "negative", "positive" or "neutral"

Sentence: Consolidated net sales increased 16 \% to reach EUR74 .8 m , while operating profit amounted to EUR0 .9 m compared to a loss of EUR0 .7 m in the prior year period . 
 
 Label:  & \textcolor{red}{NNP CD NNP CD} \\
\hline
2.&
Definition: Given a sentence do token classification on it seek to locate and classify named entities mentioned in the sentence provided. The pre-defined named entity categories along with there labels are Person (PER), Location (LOC), Organization (ORG) and Miscellaneous (MIS). If the token is not an entity mark it as None. As the entity is more than two tokens long use the prefix B with the named entity token to represent the beginning and  use the prefix I till the entity ends.

Sentence: Total shares to be offered 2.5 million 

Label: O O O O O O O

Definition: Given a sentence mined from a financial news article, you are to determine the sentiment polarity of the sentence. The task deals with financial sentiment analysis. Based on the sentiment conveyed by the sentence, label the sentence as "negative", "positive" or "neutral"

Sentence: The offer of some 30 million shares aimed to raise more than  x20ac 500 million US\$ 640 million , was expected to be completed by Oct. 9 , Outokumpu said . 

Label:

& \textcolor{red}{N N N N N N N} \\
\hline
3.&Definition: Given a question answering task from the 3rd to 9th-grade science exam. The question contains four options "A.", "B.", "C." and "D." Select the most appropriate choice that answers the question

Question: Which object in our solar system reflects light and is a satellite that orbits around one planet?

A. Earth

B. Mercury

C. the Sun

D. the Moon 

Answer: D

Definition: Given a question answering task from the 3rd to 9th-grade science exam. The question contains four options "A.", "B.", "C." and "D." Select the most appropriate choice that answers the question

Question: Which of the following sets contains only objects that shine as a result of reflected light?

A. moons, planets, and comets

B. moons, comets, and stars

C. planets, stars, and comets

D. planets, stars, and moons 

Answer: & \textcolor{red}{D} \\
\hline
4.&Definition: Given a context and a question do binary true and false type text classification. You are given a passage as context and a question related to the passage that can be answered as "True" or "False". Based on the context, question and your reasoning ability answer in a "True" and "False".

Context: A blacklight (or often black light), also referred to as a UV-A light, Wood's lamp, or simply ultraviolet light, is a lamp that emits long-wave (UV-A) ultraviolet light and not much visible light. 

Question: is a black light and an ultraviolet light the same thing 

Answer: True

Definition: Given a question from a scientific exam about Physics, Chemistry, and Biology, among others. The question is in multiple choice format with four answer options "A.", "B.", "C." and "D.". Using your knowledge about the scientific fields answer the question and provide the label "A", "B", "C" and "D" as answer

Question: When exposed to ultraviolet, some substances, such as minerals, glow in characteristic visible wavelengths, a process called this?

A. pigment

B. plasma

C. chemical reaction

D. fluorescence 

Answer: & \textcolor{red}{fluorescence}\\
\hline
\end{tabular}
}
\caption{Error analysis of cross-task prompting. Four examples represent the major error characteristics. The outputs shown are generated by LLaMA-2 13B.}
\vspace{-4mm}
\label{tab:error-analysis-complete}
\end{center}
\end{table*}

%%%%%%%%%%%%%%%%%%%%%%%%%%%%%%%%%%%%%%%%%%%%%%%%%%%%%%%%%%%%%% LAYER-WISE CORRELATION %%%%%%%%%%%%%%%%%%%%%%%%%%%%%%%%%%%%%%%%%%%%%%%%%%%%%%%%%%%%%%

\section{Activation analysis on LLaMA-2 7B}
\if 0
\begin{figure*}[!t]
\begin{center}
\includegraphics[width=\linewidth]{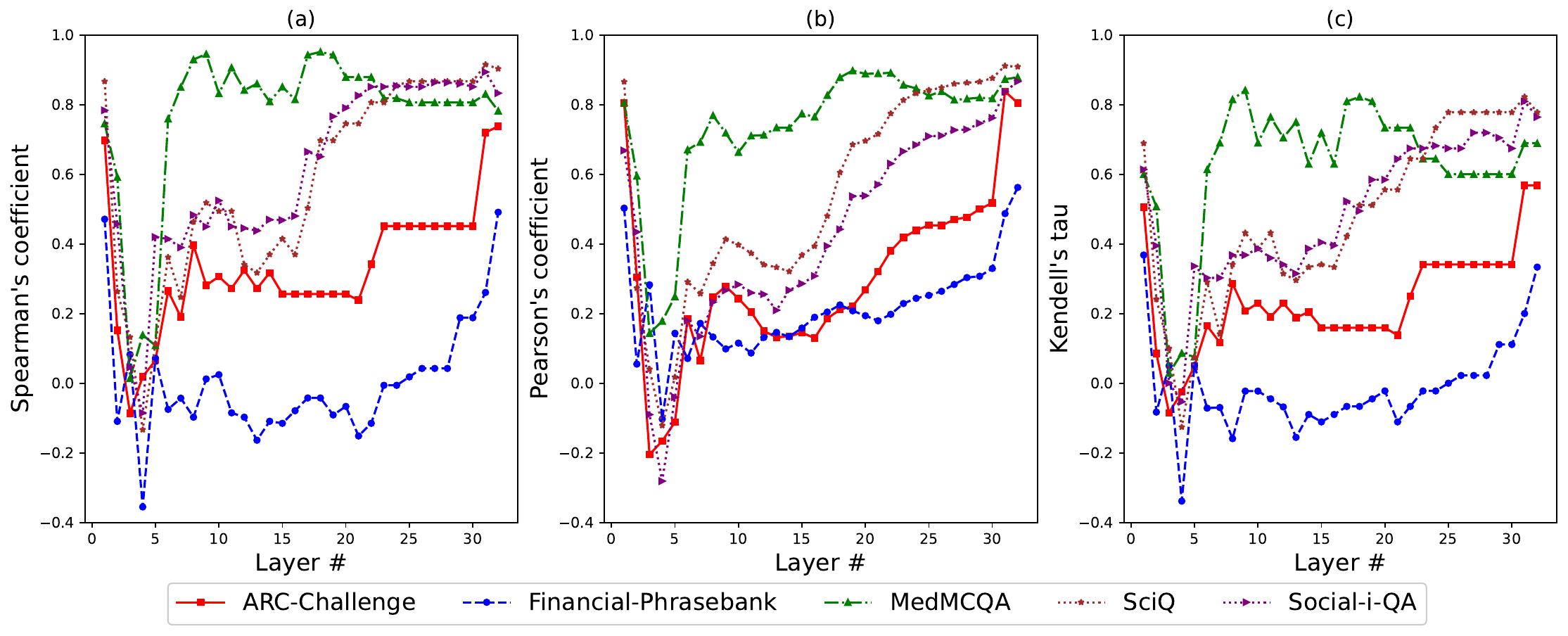}
\captionsetup{labelfont={color=black},font={color=black}}
\caption{Variation of (a) Spearman's correlation coefficient, (b) Pearson correlation coefficient and (c) Kendall’s tau, with layers as reported in Table \ref{tab:layer_correlation} for LLaMA-2 7B.}
\label{fig:corr_plot}
\end{center}
\vspace{-7mm}
\end{figure*}
\fi

Full correlation analysis is presented in Table~\ref{tab:layer_correlation}. %and Figure~\ref{fig:corr_plot}.

\begin{table*}[!t]
\small
\centering
\adjustbox{max width=1\linewidth}{
\begin{tabular}{l|ccc|ccc|ccc|ccc|ccc}
\hline
\backslashbox{Layer \#}{TAR}  & \multicolumn{3}{c}{ARC-Challenge} \vline & \multicolumn{3}{c}{Financial-Phrasebank} \vline & \multicolumn{3}{c}{MedMCQA} \vline & \multicolumn{3}{c}{SciQ} \vline & \multicolumn{3}{c}{Social-i-QA} \\ \hline
%\multicolumn{16}{c}{\cellcolor[HTML]{C0C0C0}LLaMA-2 7B} \\ \hline
\rowcolor[HTML]{C0C0C0}
 & ${\rho}_s$ & $\rho$ & $\tau$ & ${\rho}_s$ & $\rho$ & $\tau$ & ${\rho}_s$ & $\rho$ & $\tau$ & ${\rho}_s$ & $\rho$ & $\tau$ & ${\rho}_s$ & $\rho$ & $\tau$ \\
\hline
Layer-1 & 0.70 & 0.80 & 0.51 & 0.47 & 0.50 & \textbf{0.37} & 0.75 & 0.81 & 0.60 & 0.87 & 0.87 & 0.69 & 0.78 & 0.67 & 0.61  \\
Layer-2 & 0.15 & 0.30 & 0.08 & -0.11 & 0.05 & -0.08 & 0.59 & 0.59 & 0.51 & 0.26 & 0.27 & 0.24 & 0.46 & 0.43 & 0.39 \\
Layer-3 & -0.09 & -0.20 & -0.08 & 0.08 & 0.28 & 0.05 & 0.01 & 0.14 & 0.03 & 0.13 & 0.04 & 0.10 & 0.05 & -0.09 & 0.00 \\
Layer-4 & 0.02 & -0.17 & -0.03 & -0.36 & -0.10 & -0.34 & 0.14 & 0.18 & 0.09 & -0.13 & -0.12 & -0.13 & -0.09 & -0.28 & -0.05  \\
Layer-5 & 0.07 & -0.11 & 0.05 & 0.07 & 0.14 & 0.05 & 0.11 & 0.25 & 0.08  & 0.11 & 0.02 & 0.07 & 0.42 & -0.04 & 0.34 \\
Layer-6 & 0.26 & 0.19 & 0.16 & -0.08 & 0.07 & -0.07 & 0.76 & 0.67 & 0.61 & 0.36 & 0.29 & 0.29 & 0.41 & 0.18 & 0.30  \\ 
Layer-7 & 0.19 & 0.06 & 0.12 & -0.04 & 0.17 & -0.07 & 0.85 & 0.69 & 0.69 & 0.25 & 0.26 & 0.14 & 0.39 & 0.13 & 0.30 \\ 
Layer-8 & 0.40 & 0.25 & 0.29 & -0.10 & 0.13 & -0.16 & 0.93 & 0.77 & 0.81 & 0.46 & 0.34 & 0.34 & 0.48 & 0.23 & 0.37 \\ 
Layer-9 & 0.28 & 0.28 & 0.21 & 0.01 & 0.10 & -0.02 & \textbf{0.95} & 0.72 & \textbf{0.84} & 0.52 & 0.41 & 0.43 & 0.45 & 0.27 & 0.37 \\
Layer-10 & 0.31 & 0.24 & 0.23 & 0.02 & 0.12 & -0.02 & 0.83 & 0.66 & 0.69 & 0.49 & 0.40 & 0.39 & 0.52 & 0.28 & 0.39  \\
Layer-11 & 0.27 & 0.20 & 0.19 & -0.09 & 0.09 & -0.04 & 0.91 & 0.71 & 0.76 & 0.49 & 0.37 & 0.43 & 0.45 & 0.26 & 0.36 \\
Layer-12 & 0.32 & 0.15 & 0.23 & -0.10 & 0.13 & -0.07 & 0.84 & 0.71 & 0.70 & 0.34 & 0.34 & 0.31 & 0.45 & 0.26 & 0.34 \\
Layer-13 & 0.27 & 0.13 & 0.19 & -0.16 & 0.15 & -0.16 & 0.86 & 0.73 & 0.75 & 0.32 & 0.33 & 0.30 & 0.44 & 0.21 & 0.31 \\
Layer-14 & 0.32 & 0.13 & 0.20 & -0.11 & 0.13 & -0.09 & 0.81 & 0.73 & 0.63 & 0.37 & 0.32 & 0.33 & 0.47 & 0.27 & 0.39  \\
Layer-15 & 0.26 & 0.15 & 0.16 & -0.12 & 0.16 & -0.11 & 0.85 & 0.77 & 0.72 & 0.41 & 0.37 & 0.34 & 0.47 & 0.29 & 0.40  \\
Layer-16 & 0.26 & 0.13 & 0.16 & -0.08 & 0.19 & -0.09 & 0.81 & 0.76 & 0.63 & 0.37 & 0.39 & 0.33 & 0.48 & 0.31 & 0.40 \\
Layer-17 & 0.26 & 0.19 & 0.16 & -0.04 & 0.20 & -0.07 & 0.94 & 0.83 & 0.81 & 0.50 & 0.48 & 0.42 & 0.66 & 0.39 & 0.52 \\
Layer-18 & 0.26 & 0.21 & 0.16 & -0.04 & 0.22 & -0.07 & \textbf{0.95} & 0.88 & 0.82 & 0.70 & 0.61 & 0.51 & 0.65 & 0.44 & 0.49  \\
Layer-19 & 0.26 & 0.22 & 0.16 & -0.09 & 0.21 & -0.04 & 0.94 & \textbf{0.90} & 0.81 & 0.70 & 0.69 & 0.51 & 0.77 & 0.54 & 0.58  \\
Layer-20 & 0.26 & 0.27 & 0.16 & -0.07 & 0.19 & -0.02 & 0.88 & 0.89 & 0.73 & 0.75 & 0.70 & 0.56 & 0.79 & 0.54 & 0.58 \\
Layer-21 & 0.24 & 0.32 & 0.14 & -0.15 & 0.18 & -0.11 & 0.88 & 0.89 & 0.73 & 0.75 & 0.72 & 0.56 & 0.83 & 0.57 & 0.64  \\
Layer-22 & 0.34 & 0.38 & 0.25 & -0.12 & 0.20 & -0.07 & 0.88 & 0.89 & 0.73 & 0.81 & 0.77 & 0.64 & 0.85 & 0.63 & 0.67  \\
Layer-23 & 0.45 & 0.42 & 0.34 & -0.01 & 0.23 & -0.02 & 0.82 & 0.86 & 0.64 & 0.81 & 0.81 & 0.64 & 0.85 & 0.67 & 0.67  \\
Layer-24 & 0.45 & 0.44 & 0.34 & -0.01 & 0.24 & -0.02 & 0.82 & 0.85 & 0.64 & 0.85 & 0.83 & 0.73 & 0.85 & 0.68 & 0.68  \\
Layer-25 & 0.45 & 0.45 & 0.34 & 0.02 & 0.25 & 0.00 & 0.81 & 0.83 & 0.60 & 0.87 & 0.84 & 0.78 & 0.85 & 0.71 & 0.67  \\
Layer-26 & 0.45 & 0.45 & 0.34 & 0.04 & 0.26 & 0.02 & 0.81 & 0.84 & 0.60 &  0.87 & 0.85 & 0.78 & 0.85 & 0.71 & 0.67  \\
Layer-27 & 0.45 & 0.47 & 0.34 & 0.04 & 0.28 & 0.02 & 0.81 & 0.81 & 0.60 & 0.87 & 0.86 & 0.78 & 0.86 & 0.73 & 0.72 \\
Layer-28 & 0.45 & 0.48 & 0.34 & 0.04 & 0.30 & 0.02 & 0.81 & 0.82 & 0.60 & 0.87 & 0.86 & 0.78 & 0.86 & 0.73 & 0.72 \\
Layer-29 & 0.45 & 0.50 & 0.34 & 0.19 & 0.31 & 0.11 & 0.81 & 0.82 & 0.60 & 0.87 & 0.87 & 0.78 & 0.86 & 0.75 & 0.70 \\
Layer-30 & 0.45 & 0.52 & 0.34 & 0.19 & 0.33 & 0.11 & 0.81 & 0.82 & 0.60 & 0.87 & 0.88 & 0.78 & 0.85 & 0.76 & 0.67  \\
Layer-31 & 0.72 & \textbf{0.84} & \textbf{0.57} & 0.26 & 0.49 & 0.20 & 0.83 & 0.87 & 0.69 & \textbf{0.92} & \textbf{0.91} & \textbf{0.82} & \textbf{0.89} & 0.84 & \textbf{0.81} \\
Layer-32 & \textbf{0.74} & 0.80 & \textbf{0.57} & \textbf{0.49} & \textbf{0.56} & 0.33 & 0.78 & 0.88 & 0.69 & 0.90 & \textbf{0.91} & 0.78 & 0.83 & \textbf{0.87} & 0.76 \\

\hline
                                    
\end{tabular}
}
\caption{The correlation between the ranking of the source tasks for each layer of the LLM with their actual ranking obtained based on performance for each target task. The ranking for each layer is obtained based on the cosine similarity between the mean activations of the source and target task definitions derived from that layer. ${\rho}_s$, $\rho$ and $\tau$ are the \textit{Spearman rank-order correlation coefficient}, \textit{Pearson correlation coefficient} and \textit{Kendall’s tau} measures respectively. The results shown in this table are for LLaMA-2 7B.}
\label{tab:layer_correlation}
\vspace{-3mm}
\end{table*}

\begin{table*}[!t]
\tiny
\centering
\adjustbox{max width=1\linewidth}{
{\color{black}\begin{tabular}{lccc}
\hline
 SRC & p-value & t-statistic & Significant/Not \\ \hline
\multicolumn{4}{c}{\cellcolor[HTML]{C0C0C0}LLaMA-2 7B}                                                                                \\ \hline
ARC-Easy & 1.19E-102 	& 22.5127 	& Significant \\ 
AG-news & 1.62E-103 	& 22.6187 	& Significant \\ 
BoolQ 	& 1.07E-62 	& 17.1466 	& Significant \\ 
Com-QA 	& 2.20E-78 	& 19.3822 	& Significant \\ 
C-POS 	& 2.30E-14 	& 7.5864 	& Significant \\ 
C-NER 	& 2.30E-09 	& 5.8815 	& Significant \\ 
MNLI 	& 3.52E-84 	& 20.1607 	& Significant \\ 
QQP 	& 2.06E-53 	& 15.7097 	& Significant \\ 
RACE 	& 2.47E-151 & 28.079 	& Significant \\ 
SST2 	& 2.48E-90 	& 20.9642 	& Significant \\ \hline

\multicolumn{4}{c}{\cellcolor[HTML]{C0C0C0}LLaMA-2 13B}                                                                               \\ \hline

ARC-Easy 	& 2.79E-48 	& 14.8683 	& Significant \\ 
AG-news 	& 1.62E-31 	& 11.7801 	& Significant \\ 
BoolQ 	& 4.19E-29 	& 11.2767 	& Significant \\
Com-QA 	& 3.18E-36 	& 12.7125 	& Significant \\ 
C-POS 	& 6.77E-05 	& -3.8223 	& NOT Significant \\ 
C-NER 	& 7.71E-02 	& -1.4253 	& NOT Significant \\ 
MNLI 	& 3.12E-27 	& 10.8716 	& Significant \\ 
QQP 	& 6.46E-30 	& 11.4484 	& Significant \\ 
RACE 	& 2.62E-47 	& 14.7043 	& Significant \\ 
SST2 	& 2.94E-21 	& 9.4765 	& Significant \\ \hline

\multicolumn{4}{c}{\cellcolor[HTML]{C0C0C0}GPT3.5}                                                                               \\ \hline
ARC-Easy 	& 2.73E-13 	& 7.2515 	& Significant \\ 
AG-news 	& 9.83E-12 	& 6.7395 	& Significant \\ 
BoolQ 	& 5.04E-11 	& 6.4936 	& Significant \\ 
Com-QA 	& 1.82E-10 	& 6.2939 	& Significant \\ 
C-POS 	& 6.12E-09 	& -5.7154 	& NOT Significant \\ 
C-NER 	& 2.07E-04 	& -3.5359 	& NOT Significant \\ 
MNLI 	& 2.39E-03 	& 2.8244 	& Significant \\ 
QQP 	& 4.05E-04 	& 3.3535 	& Significant \\ 
RACE 	& 8.00E-10 	& 6.0566 	& Significant \\ 
SST2 	& 7.04E-04 	& 3.1968 	& Significant \\ \hline
\end{tabular}}
}
\captionsetup{labelfont={color=black},font={color=black}}
\caption{p-value and t-statistic obtained on performing a one-tailed T-test for each source task over all samples of the five target tasks. Significance is decided by taking $\alpha = 0.05$ (Abbreviations: Com-QA \(\rightarrow\) Commonsense-QA, C-POS \(\rightarrow\) Conll2003-POS, C-NER \(\rightarrow\) Conll2003-NER).}
\label{tab:significance-test}
\vspace{-4mm}
\end{table*}

\end{document}